\documentclass[10pt,twocolumn,letterpaper]{article}

\usepackage{iccv}
\usepackage{times}

\usepackage{graphicx}
\usepackage{amsmath}
\usepackage{amssymb}
\usepackage{booktabs} 
\usepackage{caption}
\usepackage{array}
\usepackage{subcaption}

\usepackage{floatrow}
\newfloatcommand{capbtabbox}{table}[][0.7\linewidth] 

\usepackage{color}
\usepackage[pagebackref,breaklinks,colorlinks]{hyperref}
\iccvfinalcopy 

\def\iccvPaperID{5561} 

\ificcvfinal\fi

\usepackage[capitalize]{cleveref}
\crefname{section}{Sec.}{Secs.}
\Crefname{section}{Section}{Sections}
\Crefname{table}{Table}{Tables}
\crefname{table}{Tab.}{Tabs.}

\newcommand{\bx}{\mathbf{x}}

\newcommand{\bI}{\mathbf{I}}

\newcommand{\bV}{\mathbf{V}}

\newcommand{\bv}{\mathbf{v}}

\newcommand{\bz}{\mathbf{z}}

\newcommand{\bff}{\mathbf{f}}
\newcommand{\bF}{\mathbf{F}}

\newcommand{\bc}{\mathbf{c}}

\newcommand{\btheta}{\boldsymbol{\theta}}

\newcommand{\bbeta}{\boldsymbol{\beta}}

\newcommand{\bxi}{\boldsymbol{\xi}}
\newcommand{\nR}{\mathbb{R}}

\newcommand{\nE}{\mathbb{E}}

\newcommand{\cD}{\mathcal{D}}
\newcommand{\cN}{\mathcal{N}}

\newcommand{\cL}{\mathcal{L}}

\makeatletter
\DeclareRobustCommand\onedot{\futurelet\@let@token\@onedot}
\def\@onedot{\ifx\@let@token.\else.\null\fi\xspace}
\def\eg{e.g\onedot} 
\def\ie{i.e\onedot}

\def\wrt{wrt\onedot}

\def\Fig{Fig\onedot}   
\makeatother

\newcommand{\figref}[1]{\Fig~\ref{#1}}
\newcommand{\secref}[1]{Section~\ref{#1}}

\renewcommand{\eqref}[1]{Eq.~\ref{#1}}
\newcommand{\tabref}[1]{Table~\ref{#1}}

\newcommand{\boldparagraph}[1]{\vspace{0.2cm}\noindent{\bf #1:} }

\newif\ifcomment
\commenttrue
\ifcomment
	\newcommand{\ag}[1]{ \noindent {\color{red} {\bf Andreas:} {#1}} }
	\newcommand{\yl}[1]{ \noindent {\color{cyan} {\bf Yiyi:} {#1}} }

\else
	\newcommand{\ag}[1]{}
	\newcommand{\yl}[1]{}
\fi

\newcommand{\METHOD}{VeRi3D }

\newcolumntype{P}[1]{>{\centering\arraybackslash}m{#1}}

\begin{document}

\title{VeRi3D: Generative Vertex-based Radiance Fields for \\ 3D Controllable Human Image Synthesis}

\author{
Xinya Chen$^{1}$,
~Jiaxin Huang$^{1}$,
~Yanrui Bin$^{2}$,
~Lu Yu$^{1}$,
~Yiyi Liao$^{1}$\thanks{Corresponding author.} 
\vspace{0.8em}\\
$^{1}$Zhejiang University~$^{2}$Huazhong University of Science and Technology\\
{\tt\small \{hust.xinyachen,~jaceyh919,~binyanrui\}@gmail.com, \{yul,~yiyi.liao\}@zju.edu.cn}
}


\vspace{-0.2cm}
\twocolumn[{%
\renewcommand\twocolumn[1][]{#1}%
\maketitle
\thispagestyle{empty}

\begin{center}
    \vspace{-0.2cm}
    \centering
    \captionsetup{type=figure}
   \includegraphics[width=0.82\linewidth]{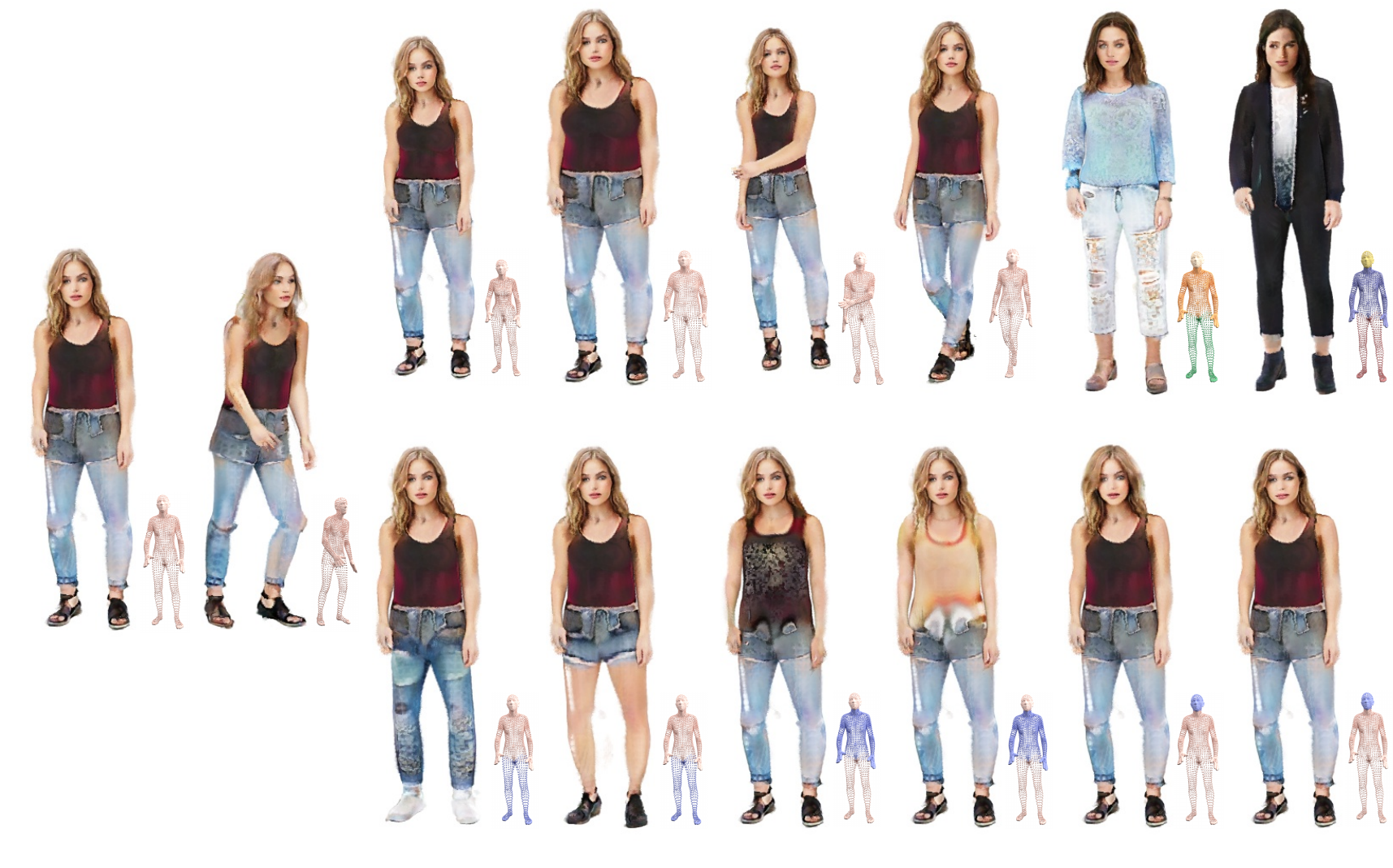}
   \vspace{-0.3cm}
    \captionof{figure}{\textbf{3D Controllable Human Image Synthesis}. Our VeRi3D enables control over shape, pose, and appearance in the first row and part-level editing of the lower body,  upper body, and head in the second row. The bottom right illustrates the vertices used to control the image synthesis, where colored vertices indicate the edited part.
    }
\end{center}%
}]
\renewcommand{\thefootnote}{\fnsymbol{footnote}}
\footnotetext[1]{Corresponding authors.}

\begin{abstract}
\vspace{-0.2cm}
    Unsupervised learning of 3D-aware generative adversarial networks has lately made much progress. Some recent work demonstrates promising results of learning human generative models using neural articulated radiance fields, yet their generalization ability and controllability lag behind parametric human models, i.e., they do not perform well when generalizing to novel pose/shape and are not part controllable. To solve these problems, we propose VeRi3D, a generative human vertex-based radiance field parameterized by vertices of the parametric human template, SMPL. We map each 3D point to the local coordinate system defined on its neighboring vertices, and use the corresponding vertex feature and local coordinates for mapping it to color and density values. We demonstrate that our simple approach allows for generating photorealistic human images with free control over camera pose, human pose, shape, as well as enabling part-level editing. Project page: \urlstyle{same} \url{https://XDimlab.github.io/VeRi3d}.
\end{abstract}
\section{Introduction}
\label{sec:intro}
Generating diverse photorealistic renderings of clothed humans has a wide range of applications including visual effects, virtual try-on, VR/AR, and creative image editing. 
Compared to designing 3D human avatars manually which is expensive and time-consuming, learning generative human models from data is a promising alternative to reduce the design effort.
To meet the requirements of the aforementioned applications, the generative model should be capable of rendering photorealistic clothed humans with free control over camera pose, human pose, shape, and appearance, ideally at the part-level, such as changing a hoody from a sweater for the same individual.



Existing generative human models do not fulfill these requirements yet. Parametric 3D body models~\cite{Anguelov2015SIGGRAPH,Loper2015SIGGRAPH,Pavlakos2019CVPR}, \eg, SMPL, faithfully capture the human pose and shape statistics with explicit control over these attributes but do not model clothing. Based on SMPL, generative models for clothed human avatars have been proposed~\cite{Chen2022CVPR,Corona2021CVPR,Ma2021ICCV}. 
However, these methods do not model appearance and require 3D supervision for modeling geometry.


Recently, 3D-aware generative adversarial networks (GANs) have shown great success in learning 3D representations, \eg, neural radiance fields, from single-view 2D supervisions.
While most 3D-aware GANs focus on non-articulated objects, e.g., faces and cars~\cite{Schwarz2020NeurIPS,Chan2020CVPR,Chan2021ARXIV,Gu2021ARXIV},
there are some recent attempts to learn generative human radiance fields from 2D images~\cite{Noguchi2022ECCV,Bergman2022NEURIPS,Zhang2022ARXIV}.
All these methods map a 3D point in the observation space to a canonical space, allowing for learning shape and appearance in a pose-agnostic space. One line of work models the mapping function via a learned blending field~\cite{Noguchi2022ECCV}, yet learning the distribution of the blending field from single-view 2D images is highly challenging and struggles to generalize to out-of-distribution poses. Another line of works model articulation using a fixed surface-based mapping without learning parameters~\cite{Bergman2022NEURIPS,Zhang2022ARXIV}. This indeed simplifies the task of the generator and leads to better generalization, but the surface-based mapping can be inaccurate, \eg, an unoccupied background point in the observation space might be mapped to an occupied foreground point in the 
canonical space and thus yielding ghosting artifacts. 
Further, despite providing control over the human pose, the controllability over the shape and fine-grained parts is yet to be explored. 

In this work, we propose generative \emph{Ve}rtex-based human \emph{R}ad\emph{i}ance fields, \METHOD, aiming for bridging the full controllability of parametric models and the image fidelity of 3D-aware GANs. 
Our key idea is to parametrize the generative human radiance fields using a set of vertices predefined by a human model, SMPL, and learning distributions over the vertex features to generative diverse human images.
Specifically, we transform a 3D point to the individual local coordinate systems of its nearest vertices, mapping the point to a color and a density value based on the feature vectors attached to the nearest vertices and its locations in the local coordinates.
Our representation has the following advantages:
i) This formulation allows us to enjoy the benefit of using a fixed mapping guided by SMPL pose, leaving the generator to learn a pose-agnostic human representation only. 
ii) Further, our formulation does not suffer from ghosting artifacts as each point is mapped to an \textit{individual} local coordinate system instead of a \textit{shared} global canonical space as in previous methods~\cite{Bergman2022NEURIPS,Zhang2022ARXIV}.
iii) Our formulation is naturally suited for generative models as we can learn a set of pose-agnostic vertex features based on a fixed UV mapping of the vertices using a 2D convolution neural network.
iv) We demonstrate that our method naturally enables control over pose and shape by editing the vertex locations. Further, it can easily achieve part-level control by manipulating vertex features of the same part.

In conclusion, our contributions are as follows:
1) We present a 3D-aware GAN method for generating controllable radiance fields of human bodies.
2) Our method introduces a human radiance field representation that enables generalization to novel poses and body shapes, as well as editing of local cloth shape and appearance. 
3) We demonstrate high-quality results for unconditional generation and animation of human bodies using several datasets, including Surreal, AIST++ and DeepFashion.

\section{Related Work}
\boldparagraph{GAN-based Image Synthesis}
Generative adversarial networks~\cite{Goodfellow2014NIPS} have achieved 2D image synthesis with high visual fidelity \cite{Mescheder2018ICML,Karras2018ICLR,Karras2019CVPR,Karras2020CVPR,Karras2021NEURIPS,Sauer2022SIGGRAPH,yang20223dhumangan}. As these unconditional methods have no explicit control over the generated images, a line of works proposes conditional human image synthesis by transferring a 2D human pose~\cite{Ma2017NeurIPS,Siarohin2018CVPR,Esser2018CVPR,Dong2018NeurIPS,Tang2020ECCV,Zhu2022PAMI,Fruhstuck2022CVPR,lwb2019}, a semantic segmentation map~\cite{Lassner2017ICCV} or a UV feature map~\cite{Grigorev2021CVPR,sarkar2021humangan} to an RGB image.
Despite enabling controllability, the shape and appearance consistency given different input conditions is not guaranteed due to the 2D-to-2D generation.
In contrast to these 2D image synthesis methods, we focus on learning to generate 3D representations that can be rendered to novel viewpoints. 

\boldparagraph{3D-Aware Image Synthesis}
Recently, 3D-aware image synthesis has attracted growing attention by lifting the generator to the 3D space. A key to 3D-aware image synthesis is the choice of the underlying 3D representation. Early methods attempt to learn 3D voxel grid~\cite{Nguyen-Phuoc2019ICCV, Henzler2019ICCV} and mesh~\cite{Liao2020CVPR}, yet these discretized representations limit the image fidelity. More recent methods exploit neural radiance fields~\cite{Mildenhall2020ECCV} for 3D-aware image synthesis ~\cite{Schwarz2020NeurIPS,Chan2020CVPR,Chan2021ARXIV,Gu2021ARXIV,Jo2021ARXIV,Xu2021ARXIV,Zhou2021ARXIV,OrEl2021ARXIV,Xu2021NEURIPS,Pan2021NEURIPS,DENG2021ARXIV,Schwarz2022NEURIPS}. Albeit achieving photorealistic 3D-aware image synthesis, these methods focus on non-articulated objects, \eg, faces and cars.

\boldparagraph{3D-Aware Human Image Synthesis}
There are a few attempts to address the task of 3D-aware human image synthesis leveraging neural radiance fields~\cite{Zhang2022ECCV,Noguchi2022ECCV,Bergman2022NEURIPS,Zhang2022ARXIV,hong2023evad}. 3D-SGAN~\cite{Zhang2022ECCV} leverages a semantic radiance field to render a semantic map,  combined with a 2D convolutional network for converting the semantic label to a textured image. Despite demonstrating photorealistic human image synthesis, the 2D texture generator hinders multi-view consistency. Our work is more related to ENARF~\cite{Noguchi2022ECCV} that generates a human radiance field in a canonical space and obtains textured 2D images via volume rendering. ENARF proposes to learn a blend weight field to map humans of different poses to a canonical space. However, this learned weight field does not generalize well to unseen poses. GNARF~\cite{Bergman2022NEURIPS} and AvatarGen~\cite{Zhang2022ARXIV} use surface-based mapping that enables better generalization to novel poses. 
Our experimental results show that surface-based mapping may suffer from ghosting artifacts and our vertex-based representation instead avoids such a problem.
More importantly, in contrast to these methods~\cite{Noguchi2022ECCV,Bergman2022NEURIPS,Zhang2022ARXIV} that all leverage the tri-plane-based representation, we propose to learn features directly on the human vertices and learn a generative vertex-based radiance field. This enables generalization to novel poses and shapes. A more recent work~\cite{hong2023evad} also enables control over the shape and pose using a compositional neural radiance field. In contrast to these methods, our method additionally enables part-level editing for free.

\boldparagraph{Articulated Human Representations}
The performance of 3D-aware human generative models is closely related to the choice of articulated human representation. A line of works represents the human body as a mesh controlled by a set of pose and shape parameters~\cite{Loper2015SIGGRAPHASIA}. While presenting compact representations, they neglect human appearance and cloth. More recently, coordinate-based MLPs have advanced many fields, including reconstruction~\cite{Mescheder2019CVPR,Park2019CVPR,Oechsle2021ICCV} and novel view synthesis~\cite{mildenhall2020nerf,Xu2022CVPR}. There are many attempts to leverage implicit neural representations for human reconstruction~\cite{Peng2021CVPR,Peng2021ICCV,Weng2022CVPR,Zhao2022CVPR,Chen2021ICCV,Xu2022CVPRa,Ma2021ICCV,liu2021neural,li2022tava,noguchi2021neural}. Among these methods, earlier methods \cite{Peng2021CVPR, Ma2021ICCV} model humans in observation space, increasing the variation to be learned. Modeling humans in a global canonical space avoids this problem, yet learning a backward weight field struggles to generalize to novel poses~\cite{Peng2021ICCV,Weng2022CVPR,Zhao2022CVPR}, and learning a forward weight field is computationally expensive due to the iterative root finding~\cite{Chen2021ICCV, li2022tava}.
Our proposed vertex-based representation is more closely related to deterministic surface-based warping methods~\cite{Xu2022CVPRa,liu2021neural}.
In contrast to all aforementioned representations focusing on reconstruction, our representation enables generalization to novel poses without expensive computation and does not suffer from ghosting artifacts, hence well-suited for generative models.



\section{Method}
\begin{figure*}[t]
  \centering
  \includegraphics[width=\linewidth]{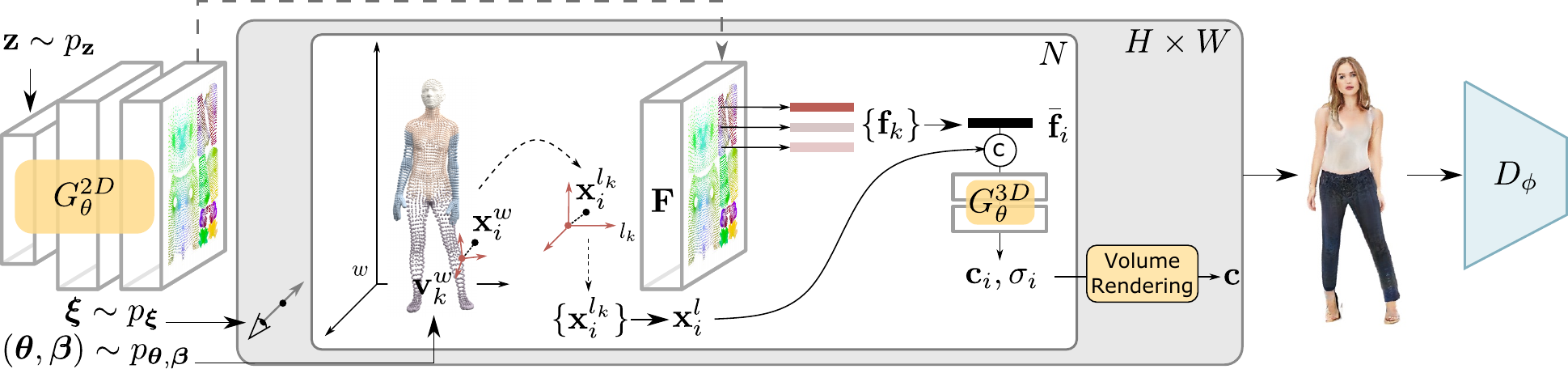}

   \caption{\textbf{Generative VeRi3D.} We model the generative human radiance fields based on a set of vertices and their corresponding feature vectors. Given a sampled camera pose $\bxi$ and sampled human pose and shape $(\btheta,\bbeta)$, we sample a 3D point $\bx_i^{w}$ along a ray in the observation space and retrieve its $K$ nearest neighbors $\{\bv_k^{w}\}_{k=1,\dots,K}$. For each neighboring vertex $\bv_k^{w}$, we transform $\bx_i^{w}$ to the local coordinate system of $\bv_k^{w}$ and obtain $\bx_i^{l_k}$. Next, we combine the weighted feature vectors $\{\bff_k\}$ and the local coordinate $\{\bx_i^{l_k}\}$ to generate the color $\bc$ and density $\sigma$ at this point. This mapping is performed on $N$ sample points on each ray and $H \times W$ pixels to obtain an output image, whereas the feature vectors are generated once using a 2D convolutional network.} 

   \label{fig:pipeline}
\end{figure*}

Given a set of 2D images, we aim to learn a generative human radiance field with explicit control over camera pose, human pose and shape, as well as enabling control over part-level shape and appearance.
In the following, we first briefly review generative radiance fields of non-articulated objects in \secref{sec:graf}. Next, we introduce our vertex-based generator in \secref{sec:smplgen} and its versatile controllability in \secref{sec:control}. Finally, we describe the training of the model in \secref{sec:training}.

\subsection{Generative Radiance Fields}
\label{sec:graf}
\boldparagraph{Generator} Generative Radiance Fields~\cite{Schwarz2020NeurIPS,Chan2020CVPR} learn a conditional neural radiance fields from a set of 2D image collections using adversarial training. Conditioned on a noise $\bz\in\nR^{M}$ sampled from a Gaussian distribution, the generator $G_\theta$ generates a radiance field that maps a 3D point $\bx$ to a corresponding RGB value $\bc$ and density value $\sigma$:
\begin{equation}
G_\theta: \nR^{L_\bx}\times\nR^{M} \rightarrow \nR^3\times\nR^+\hspace{0.3cm} (\gamma(\bx), \bz) \mapsto (\bc, \sigma)
\end{equation} 
where $\gamma(\bx)$ is the positional encoding operation that maps the given input to higher-dimensional features and $L_\bx$ denotes the dimension of $\gamma(\bx)$. Note that we do not consider the viewing direction here for simplicity as articulated humans are usually Lambertian~\cite{Chan2021ARXIV,Schwarz2022NEURIPS}.

\boldparagraph{Volume Rendering} The generative radiance field is rendered to a 2D image via the volume rendering operation. Given a camera pose $\bxi$ sampled from a camera pose distribution $p_\xi$, the generator is queried on $H\times W$ rays with $N$ sampling points on each ray. Let $\{\bc_i, \sigma_i\}_{i=1,\dots,N}$ denote the queried color and density values of one ray, the pixel color $\bc$ at the corresponding ray is obtained via volume rendering:
\begin{gather}
\pi: (\nR^{3}\times\nR^+)^N \rightarrow \nR^3 \hspace{1cm} \{(\bc_i,\sigma_i)\} \mapsto \bc \nonumber \\
\bc = \sum_{i=1}^N \, T_i \, \alpha_i \, \bc_i \hspace{0.2cm}
T_i = \prod_{j=1}^{i-1}\left(1-\alpha_j\right) \hspace{0.2cm}
\alpha_i = 1-\exp(-\sigma_i \delta_i) \nonumber
\label{eq:volume_rendering}
\end{gather}
where $T_i$ and $\alpha_i$ denote the transmittance and alpha value at a sample point $\bx_i$, $\delta_i$ denotes the distance interval between two adjacent sample points.

\boldparagraph{Challenges} Despite achieving superior performance on non-articulated objects, e.g., human faces and cars, it is non-trivial to extend generative radiance fields to articulated objects. The na\"ive approach of generating articulated humans in the observation space (e.g., EG3D) leads to unsatisfying performance and a lack of controllability. Existing attempts for generative human radiance fields map a 3D point in the observation space to a pose-agnostic canonical space, either achieved by learning a blend weight field~\cite{Noguchi2022ECCV} or using a fixed surface-based warping~\cite{Bergman2022NEURIPS,Zhang2022ARXIV}. Learning generative blend weight fields from single-view 2D images using adversarial loss is quite challenging. Further, the learned backward weight fields struggle to generalize to novel poses out of the dataset distribution. On the other hand, surface-based warping methods avoid the challenging task of learning the blend weight fields and enable better generalization. However, it may suffer from inaccurate surface mapping and thus yielding ghosting artifacts. We propose a simple approach that maps a 3D point to the local coordinate systems of its nearest vertices. Our formulation avoids learning the weight field and do not suffer from ghosting artifacts. Moreover, our formulation allows for generalizing to novel shapes and poses, as well as editing local parts. 




\subsection{\METHOD as Generative Human Fields}
\label{sec:smplgen}

\figref{fig:pipeline} gives an overview of our method. We train our generative \METHOD on 2D image collections and their corresponding camera parameters and human pose distributions using adversarial training. We refer to the space under a sampled SMPL pose as the \textit{observation space}. Given a set of vertices of SMPL, we first define a local coordinate system for each vertex conditioned on the pose.
Next, we transform a sampled 3D point in the world coordinate of the observation space to the local coordinate systems of its neighboring SMPL vertices, representing each 3D point using the vertex feature and the local coordinates. This formulation enables generalization to novel shapes and poses. 
The vertex features are generated using a 2D convolutional neural network based on a fixed UV mapping.

More formally, let $\bz$ denote a random noise vector, $(\btheta,\bbeta)$ a sampled SMPL pose and shape, and $\bxi$ a sampled camera pose.
We define the world coordinate system of the observation space as $w$. Further, let $\bV^{w} \in \nR^{M\times 3}$ denote $M$ vertices of the SMPL model in the observation space where each vertex $\bv^{w}$ is associated with a feature vector $\bff$.
Let $T_a^b \in SE(3)$ denote a rigid transformation, \ie, $\bx^b=T_a^b \bx^a$.

Given a sampled 3D point $\bx^{w}$ in the observation space, we first retrieve its $K$ nearest neighbors from the vertices  $\{\bv_k^{w} \in\cN({\bx^{w}})\}_{k=1,\cdots,K}$ and transform the 3D points to the local coordinate systems defined on $\{\bv_k^{w}\}$. Next, $\bx^{w}$ is mapped to a color and a density value based on its local coordinates determined by $\{\bv_k^{w}\}$ and their corresponding feature vectors $\{\bff_k\}$.

\boldparagraph{Vertex-Based Local Coordinate Systems}
For each vertex $\bv_k^{w}$, we define a local coordinate system $T_{l_k}^{w}$ with its origin at the vertex, allowing for transforming a 3D point in the world coordinate system to the local coordinate system via:
\begin{equation}
\bx^{l_k} = (T_{l_k}^{w})^{-1} \bx^{w} = T^{l_k}_{w} \bx^{w},
\end{equation}
%
where $l_k$ denotes the local coordinate system of the $k$th neighbor. Note that $\bx^{l_k}$ should be pose-agnostic, \eg, the color and density of a point should be invariant to the given pose. Thus, we predefine a fixed local coordinate system $T_{\bar{l}_k}^{w}$ under the T-pose and transform it to $T_{l_k}^{w}$ via linear blend skinning. 
Concretely, the translation of $T^{w}_{\bar{l}_k}$ is $\bar{\bv}_k^{w}$ defined under the T-pose. The rotation is determined by the normal vector of $\bar{\bv}_k^{w}$ and a fixed up direction $[0,0,1]^T$. The vertex normal is calculated as the weighted average of the normals of all the faces that the vertex is part of. This vertex normal is taken as the $z$-axis of the rotation matrix. Next, we define the rest two axes using a fixed vector in the world coordinate system $[0,0,1]^T$. The $x$-axis is the cross product of the $z$-axis and $[0,0,1]^T$, and the $y$-axis is the cross product of the other two axes.


%


Next, we transform $T^{w}_{\bar{l}_k}$ to $T^{w}_{l_k}$ based on linear blend skinning using the skinning weight of the vertex:

\begin{equation}
   T^{w}_{l_k} = \left(\sum_{p=1}^{P} W(\bar{\bv}^{w}_{k})_{p} G_p(\theta)\right) T^{w}_{\bar{l}_k}
\end{equation}
where $P$ is the number of parts defined in SMPL, $G_p(\theta) \in SE(3)$ is a transformation matrix of the $p$th part given pose $\theta$, and $W(\bar{\bv}^{w}_{k})_{p}$ is the blend weight of the $p$th part. 

\boldparagraph{Vertex-based Radiance Fields}
Our vertex-based radiance fields map each point $\bx^{w}$ to a color and a density value based on its $K$ nearest vertices. This design is similar to Point-NeRF~\cite{Xu2022CVPR} but we focus on learning generative radiance fields instead of overfitting to a single scene or object. Concretely, after obtaining local coordinates $\{\bx^{l_k} \}_{k=1\dots K}$, we query the color and density of the point based on the feature vectors $\{\bff_k\}$ and $\{\bx^{l_k} \}$. Here, $\{\bff_k\}$ provides the information to distinguish different local coordinate systems, and $\{\bx^{l_k} \}$ further distinguishes points within the same local coordinate systems.
This brings several advantages: 1) Our local coordinates are attached to the SMPL vertices, thus naturally enabling pose and shape control; 2) This representation enables part-level control as we directly learn a set of features associated with the SMPL vertices.

More specifically, 
we simply accumulate the feature vectors of the $K$ nearest vertices as follows:
\begin{equation}
\bar{{\bff}} =  \sum_{k} {\frac{{p}_{k}}{\sum{p}_{k}}} {\bff}_{k}
\end{equation}
where ${p}_{k}$ is the inverse-distance weight widely used in scattered data interpolation. The intuition behind this is that closer vertices should contribute more to the output.

Regarding the local coordinates $\{\bx^{l_k} \}$, we average their directions and summarize the statistics of their norms to be order-invariant \wrt the $K$ nearest neighbors. The statistics contain minimum, maximum, mean, and variance of $K$ norms. We concatenate the 3-dimensional averaged directions and 4-dimensional norm statistics to a vector $\bx^l\in\nR^{7}$ to indicate the local coordinate information.


We use an MLP to map the accumulated feature vector $\bar{\bff}$ and the local coordinate $\bx^l$ to the color and RGB values:
\begin{equation}
G_\theta^{3D}: \nR^{L_F}\times\nR^{L_{\bx^l}} \rightarrow \nR^3\times\nR^+\hspace{0.3cm} (\bar{\bff}, \gamma(\bx^l)) \mapsto (\bc, \sigma)
\end{equation} 
where $\gamma(\cdot)$ denotes positional encoding.

\boldparagraph{Vertex Feature Generator} We leverage the UV mapping to project SMPL vertices $\bV \in \nR^{M\times 3}$ to a 2D map, enabling us to learn features on SMPL vertices leveraging a 2D CNN. In practice, we use a style-based 2D CNN to learn to generate a 2D feature map $\bF \in \nR^{H_F \times W_F \times L_F}$ given a sampled noise vector $\bz$:
\begin{equation}
G_\theta^{2D}: \nR^{M} \rightarrow \nR^{H_F \times W_F \times L_F} \hspace{0.5cm} \bz \mapsto \bF
\end{equation}
Note that the projected SMPL vertices may not fall on integer locations. We use bilinear interpolation to query a feature $\bff \in \nR^{L_F}$ from $\bF$ for each vertex.

\subsection{VeRi3D for Controllable Image Synthesis}
\label{sec:control}
\boldparagraph{Pose Control} Our model enables controlling poses using the pose parameter $\btheta$. Thanks to our vertex-based formulation, our method inherits the generalization ability of SMPL, thus enabling generalizing to poses out of the training images.

\boldparagraph{Shape Control}
Since our representation is shape-agnostic, we can control the body shape of generated humans by manipulating the body mesh surface. By controlling the shape parameter $\bbeta$ of SMPL, \ie, the coefficients of 10 orthogonal basis obtained by principal component analysis (PCA), we can obtain different body shapes. 

\boldparagraph{Part-level Control}
As we directly learn a set of features on SMPL vertices, we can control the part-level appearance and geometry by changing the vertices' features.
These vertices $\bV$ are segmented to different body parts based on the SMPL skinning weights. For vertices of each part, we apply PCA to the vertices features and get the principal component. Specificly, we randomly sample 1k latent code to compute PCA component and select the top 30 components with the largest eigenvalues. This enables part control by changing the principal coefficients. 

\subsection{Discriminator and GAN Training}
\label{sec:training}

\boldparagraph{Discriminator}
Following~\cite{Karras2019stylegan2}, we adopt a normal 2D convolutional neural network as the discriminator. Our discriminator is not conditioned on the camera pose or the human pose.

\boldparagraph{GAN Training} Given a latent code $\bz$, a camera pose $\bxi$, human pose and shape $(\btheta, \bbeta)$, and a real image $\bI$ sampled from the real-data distribution $p_\cD$, we train our \METHOD using non-saturated GAN loss with R1 regularization~\cite{Mescheder2018ICML}.
\begin{align}
\cL  = &
\nE_{\bz\sim \mathcal{N}(0, \mathbf{1}), \xi\sim p_\xi,(\btheta,\bbeta)\sim p_{\btheta,\bbeta}}
\left[f(D_\phi\left(G_\theta(\bz,\bxi,\btheta,\bbeta)\right))\right]\nonumber \\
&\,+\, \nE_{\bI \sim p_{\cD}}
\left[
f(-D_\phi(\bI))
\,+\, \lambda {\Vert \nabla D_\phi(\bI)\Vert}^2
\right] \nonumber 
\end{align}



\begin{figure*}[t!]
     \setlength{\tabcolsep}{0pt}
     \def\mywidth{17cm}
     \begin{tabular}{P{0.5cm}P{\mywidth}}
     \rotatebox{90}{ENARF~\cite{Noguchi2022ECCV}} &\includegraphics[width=\mywidth]{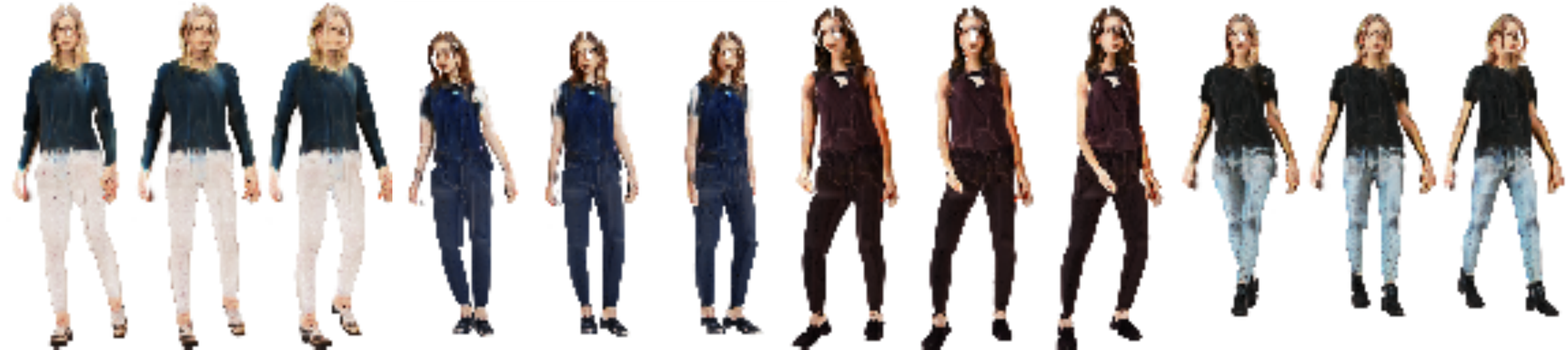} \\
    \rotatebox{90}{Tri. + Surf.} &\includegraphics[width=\mywidth]{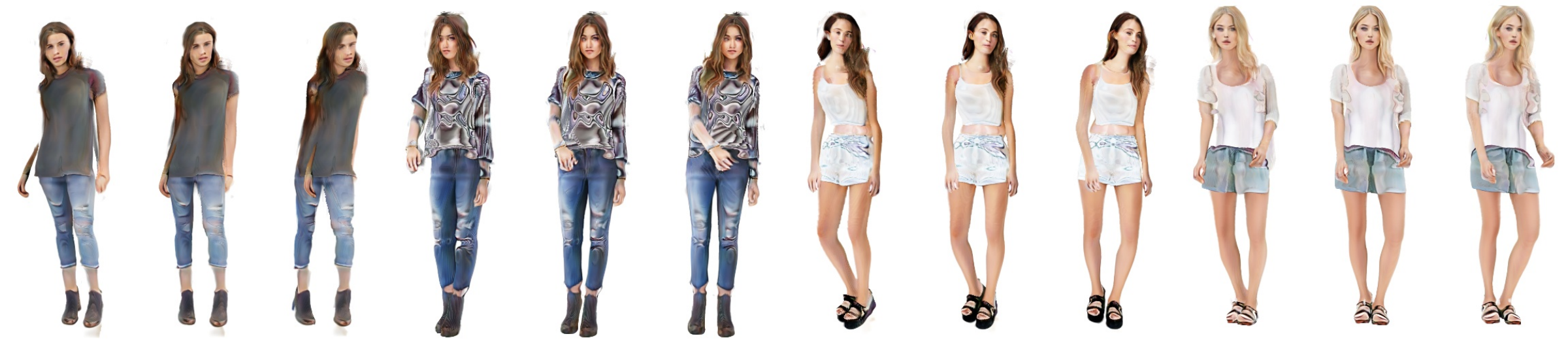} \\
     \rotatebox{90}{Ours} &\includegraphics[width=\mywidth]{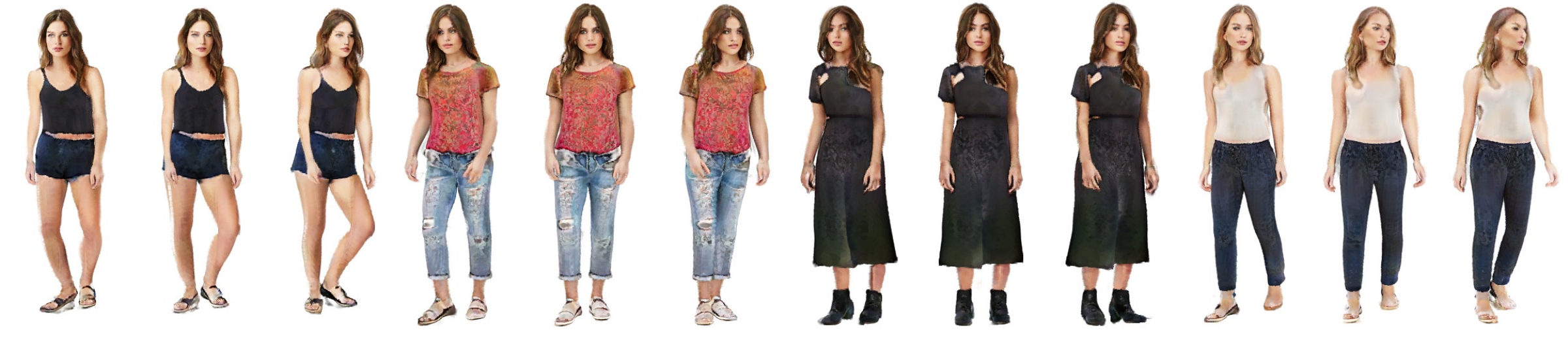}\\
     \end{tabular}\vspace{-0.2cm}
     \caption{\textbf{Qualitative Comparison on DeepFashion.} We show each sample from three viewing directions.}
     \vspace{-0.2cm}
   \label{fig:enarf_comparison}
    \end{figure*}

\begin{figure*}[t!]
    \centering
     \setlength{\tabcolsep}{0pt}
     \def\mywidth{0.1}
     \begin{tabular}{ccc@{\hskip 15pt}ccc@{\hskip 15pt}ccc}
      \includegraphics[width=\mywidth\linewidth]{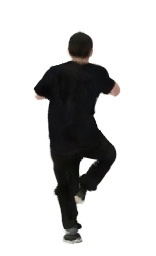} &
      \includegraphics[width=\mywidth\linewidth]{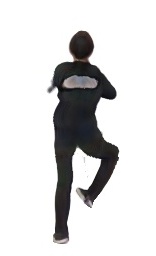} &
      \includegraphics[width=\mywidth\linewidth]{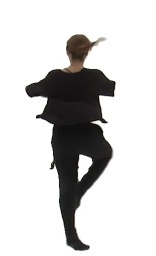} &
      \includegraphics[width=\mywidth\linewidth]{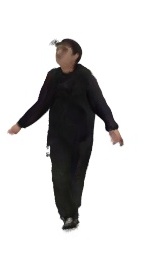} &
      \includegraphics[width=\mywidth\linewidth]{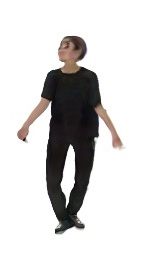} &
      \includegraphics[width=\mywidth\linewidth]{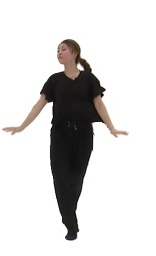} &
      \includegraphics[width=\mywidth\linewidth]{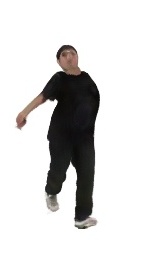} &
      \includegraphics[width=\mywidth\linewidth]{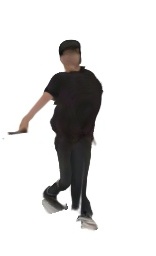} &
      \includegraphics[width=\mywidth\linewidth]{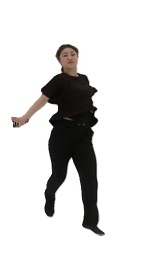}\\
      \vspace{-0.1cm}
      \begin{small}Ours\end{small}& \begin{small}Tri. + Surf.\end{small} & \begin{small}Input\end{small} &
      \begin{small}Ours\end{small}& \begin{small}Tri. + Surf.\end{small} & \begin{small}Input\end{small} &
      \begin{small}Ours\end{small}& \begin{small}Tri. + Surf.\end{small} & \begin{small}Input\end{small}\\
     \end{tabular}\vspace{-0.1cm}
     \caption{\textbf{Qualitative Comparison on AIST++}. We provide the corresponding real image as a reference for each input pose. Note that ``Tri. + Surf.'' sometimes fails to follow the input pose (1st and 2nd) and may have ghosting artifacts (3rd). }
   \label{fig:triplane_comparison}
    \end{figure*}

\begin{table*}[t]
 \small
  \centering
  \begin{tabular}{c|c|c|c|c|c|c|c|c|c}
    \toprule
     & \multicolumn{3}{c}{Deepfashion} & \multicolumn{3}{|c}{Surreal} & \multicolumn{3}{|c}{AIST++}\\
    Method & Res. & FID10k~$\downarrow$ & PCKh@0.5~$\uparrow$ & Res. & FID10k~$\downarrow$ & PCKh@0.5~$\uparrow$ & Res. & FID10k~$\downarrow$ & PCKh@0.5~$\uparrow$\\
    \midrule
    EG3D~\cite{EG3D} & 512 & 27.2 & - & 128 & 19.5 & - & 256 & 38.2 & -\\
    ENARF~\cite{Noguchi2022ECCV} & 128 & 60.3 & \textbf{0.991} & 128 & 21.3 & 0.855 & 128 & 95.7 & 0.916\\
    Tri. + Surf. & 512 & 23.7 & 0.936 & 128 & 8.0 & 0.883 & 256 & 32.4 & 0.949\\
    Ours & 512 & \textbf{21.4} & 0.988 & 128 & \textbf{6.7} & \textbf{0.984} & 256 & \textbf{32.3} & \textbf{0.973}\\
    \bottomrule
  \end{tabular}
  \caption{\textbf{Quantitative Results} on Deepfashion, Surreal, and AIST++ dataset.}
  \label{tab:GAN Results}
\end{table*}

\begin{figure*}[h]
    \vspace{-0.2cm}
     \setlength{\tabcolsep}{0pt}
     \def\mywidth{.5}
     \begin{tabular}{cc}
       \includegraphics[width=\mywidth\linewidth]{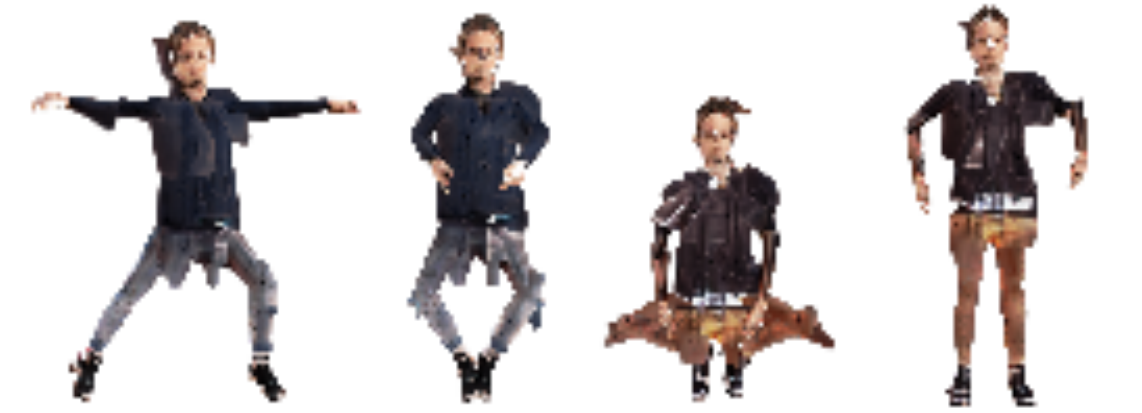} &
      \includegraphics[width=\mywidth\linewidth]{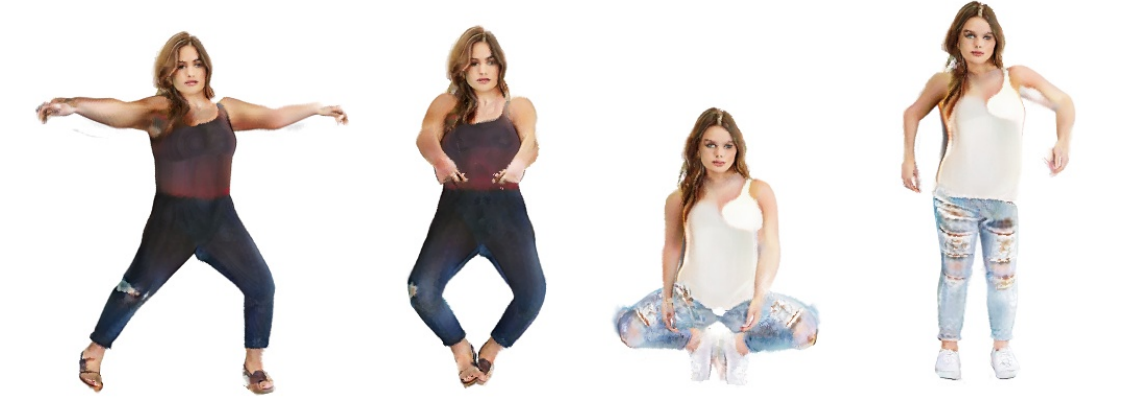} \\
      \vspace{-0.1cm}
      \begin{small}ENARF~\cite{Noguchi2022ECCV}\end{small}&
      \begin{small}Ours\end{small} \\
     \end{tabular}
     \caption{\textbf{Pose Control}. We compare pose controllability with ENARF by applying poses of the AIST++ dataset to the generative radiance fields trained on DeepFashion.}
   \label{fig:novel_pose}
    \end{figure*}

\begin{figure*}[t!]
     \setlength{\tabcolsep}{0pt}
     \def\mywidth{.31}
     \begin{tabular}{c@{\hskip 15pt}c@{\hskip 15pt}c}
     \includegraphics[width=\mywidth\linewidth]{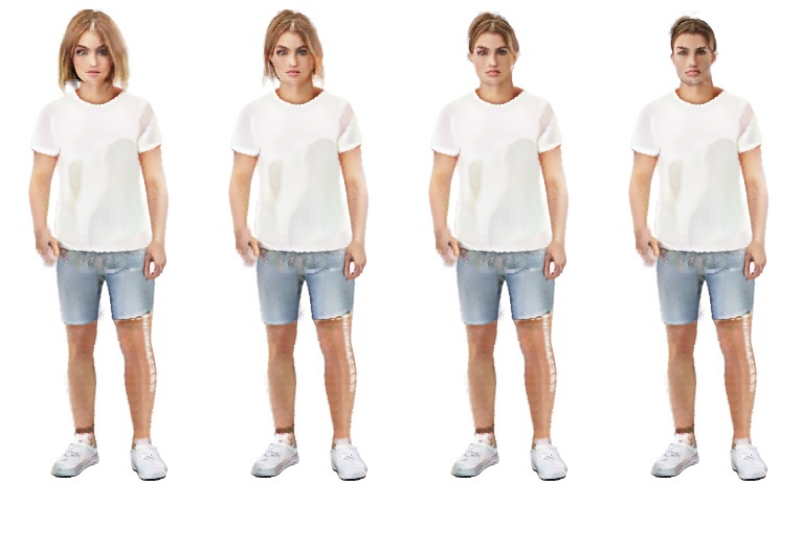} &
     \includegraphics[width=\mywidth\linewidth]{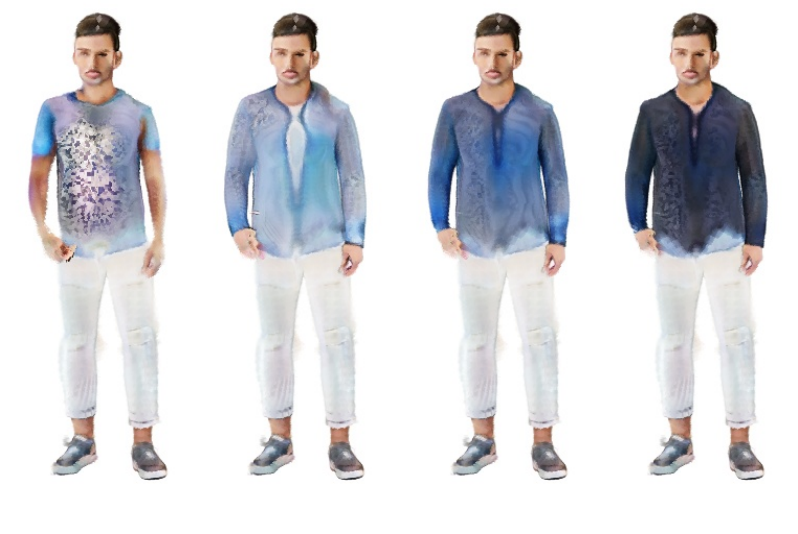}&
      \includegraphics[width=\mywidth\linewidth]{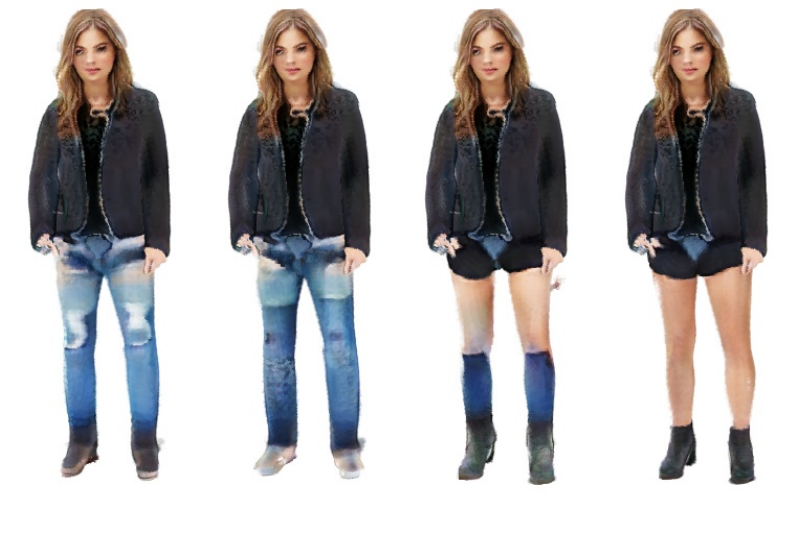} \\
      \vspace{-0.1cm}
      \begin{small}Head Control\end{small} &
      \begin{small}Upper Body Control\end{small} &
      \begin{small}Lower Body Control\end{small} \\
     \end{tabular}
     \caption{\textbf{Part Control.} Our method enables controlling the head, upper body, and lower body independently, with cloth color, style, hairstyle, and facial changes.}
   \label{fig:Part_Controllability}
    \end{figure*}

\begin{figure*}[t]
  \centering
   \includegraphics[width=0.95\linewidth]
   {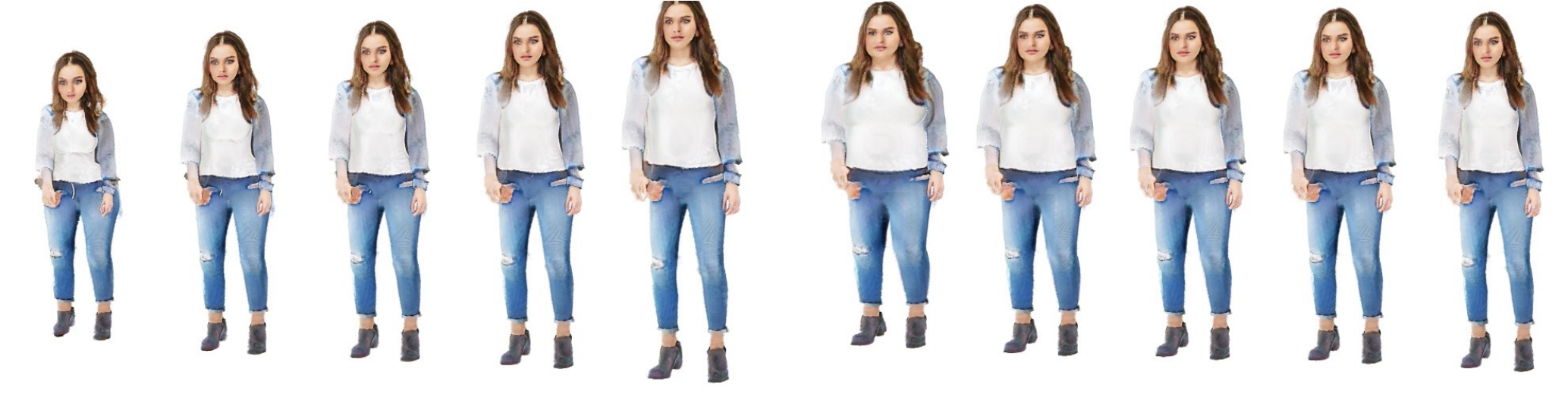}
   \caption{\textbf{Shape Control.} We manipulate the coefficiency of the first principle component for the  left four images and the second principle component for the right ones.}
   \label{fig:Shape_Controllability}
\end{figure*}


\section{3D-Aware Human Synthesis}

\boldparagraph{Datasets} 
We conduct experiments on three datasets: DeepFashion~\cite{Deepfashion}, AIST++~\cite{aist++} and Surreal~\cite{Surreal}. DeepFashion is a real-world fashion dataset. We use the data provided by~\cite{hong2023evad}. 
The dataset filters out images with partial observations and inaccurate SMPL estimations, resulting in 8K images for training. AIST++ is a video dataset capturing 30 performers in dance motion. Each frame is annotated with the ground truth camera and fitted SMPL body model. We use a segmentation model~\cite{paddleseg2019} to remove backgrounds and get 90k images in total. Surreal contains 6M images of synthetic humans created by the SMPL body model in various poses rendered in indoor scenes.


\boldparagraph{Evaluation Metrics}  
We adopt Fréchet Inception Distance (FID)~\cite{Heusel2017NIPS} to measure image quality. Following ENARF, we use the Percentage of Correct Keypoints (PCKh@0.5) to evaluate the correctness of generated poses.

\boldparagraph{Baselines} 
We compare our method against a state-of-the-art 3D-aware image synthesis model, EG3D~\cite{EG3D}, and a state-of-the-art 3D-aware human image synthesis model, ENARF~\cite{Noguchi2022ECCV}. As EG3D does not model human articulation, we provide the SMPL pose and shape parameters as input conditions to EG3D. We also compare our method with a surface-based baseline. Following AvatarGen~\cite{Zhang2022ARXIV}, we map a 3D point in the observation space to a canonical space based on SMPL skinning weights. We use a tri-plane to represent the canonical space, where we sample feature for each point in canonical space to predict density and color. We refer to this method as ``Tri. + Surf.''. We implement this baseline in our framework to ensure a fair comparison.

\boldparagraph{Implementation Details} We use a feature dimension of $256\times 256 \times 64$. The local coordinate vector $\bx^l$ has a length of $7$ which includes direction and distance statistics.  Similar to~\cite{hong2023evad}, we only sample points around the human body and increase the sampling of side view for the DeepFashion dataset. We use the Adam optimizer, with a batch size of 8 for training. The learning rate of the generator is 0.0001. The learning rate of the discriminator is 0.0002. For the DeepFashion dataset, we train 620k iterations. For the Surreal dataset, we train 640k iterations. For the AIST++ dataset, we train 400k iterations.

\subsection{3D-Aware Human Image Synthesis}

\tabref{tab:GAN Results} shows our quantitative comparison to the baselines. Note that our method achieves superior performance in terms of FID. 
While EG3D leads to reasonable FID scores, it is not controllable in terms of the human pose and thus we do not evaluate its PCKh@0.5.
Our method outperforms ENARF which maps a 3D point to the canonical space based on the learned weight fields. ENARF performs better on datasets with smaller pose variation, \ie, Surreal and DeepFashion, yet the performance degenerates on the AIST++ dataset which captures dancing videos.
Regarding the surface mapping-based approach ``Tri. + Surf.'', its FID scores are comparable to our proposed method, but the generated images are of lower accuracy in terms of pose correctness.
Compared with ENARF, our method is superior on Surreal and AIST++ and similar on the DeepFashion dataset in terms of PCKh@0.5.
Note that both ENARF and our method can maintain the input pose accurately on DeepFashion as its poses are of lower variation.

We show the qualitative comparison to ENARF and Tri. + Surf. in \figref{fig:enarf_comparison} on the DeepFashion dataset. Aligned with the quantitative results, our method produces images with higher visual fidelity compared to ENARF.
Despite achieving similar visual quality on DeepFashion, the Tri. + Surf. baseline sometimes leads to inaccurate poses and ghosting artifacts when training on datasets with larger pose variations. This can be seen in \figref{fig:triplane_comparison}, which shows our qualitative comparison to Tri. + Surf. on the AIST++ dataset.


\subsection{Controllability}
\begin{figure}[t]
  \centering
  \begin{subfigure}{0.97\linewidth}
    \includegraphics[width=0.99\linewidth]{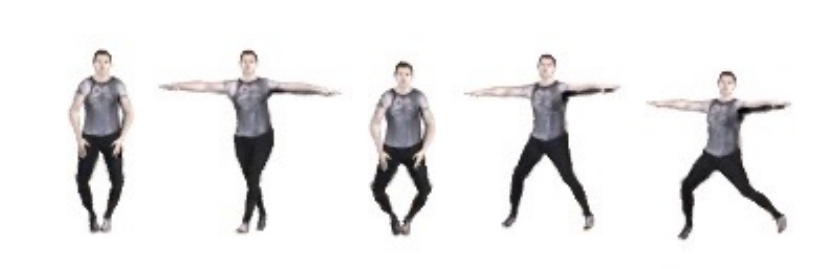}
    \vspace{-0.3cm}
    \caption{Pose Control}
  \end{subfigure}
  \begin{subfigure}{0.97\linewidth}
   \includegraphics[width=0.98\linewidth]{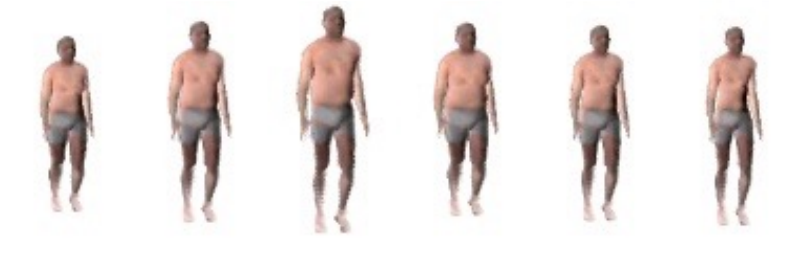}
   \vspace{-0.4cm}
    \caption{Shape Control}
  \end{subfigure}
   \caption{\textbf{Qualitative Results} on Surreal Dataset. }
   \label{fig:surreal}
\end{figure}

\begin{figure}[t]
  \centering
  \begin{subfigure}{0.97\linewidth}
    \includegraphics[width=0.99\linewidth]{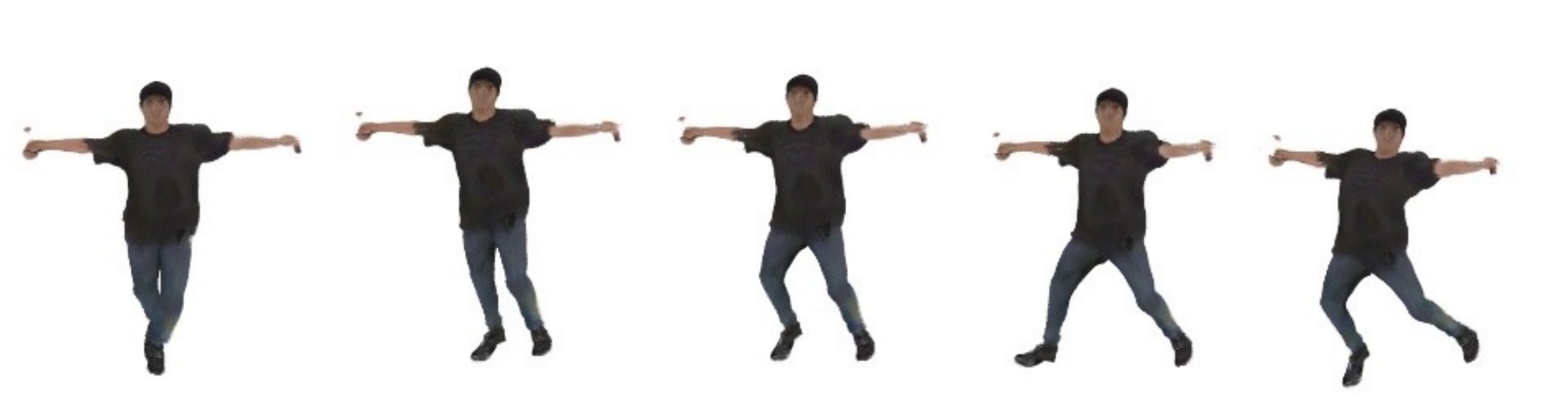}
    \vspace{-0.3cm}
    \caption{Pose Control}
  \end{subfigure}
  \begin{subfigure}{0.97\linewidth}
   \includegraphics[width=0.98\linewidth]{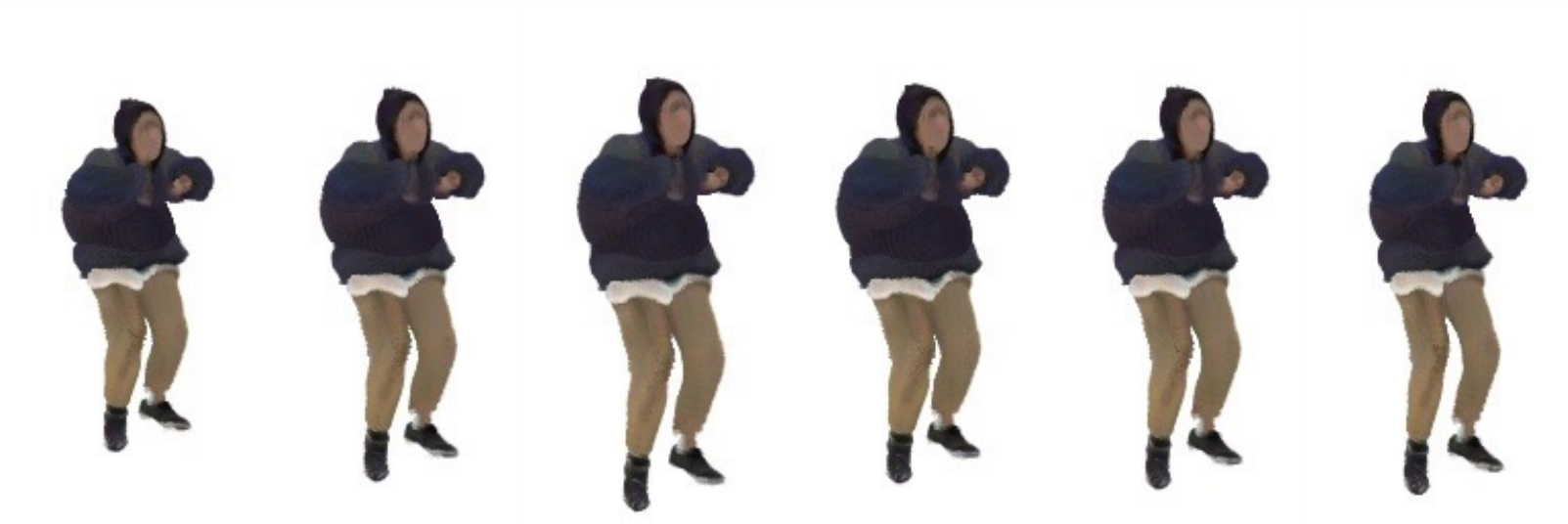}
   \vspace{-0.1cm}
    \caption{Shape Control}
  \end{subfigure}
   \caption{\textbf{Qualitative Results} on AIST++ Dataset. }
   \label{fig:aist}
\end{figure}

\boldparagraph{Pose Control}
Our vertex-based representation enables generalization to out-of-distribution novel poses. To evaluate our generalization ability, we apply the poses of the AIST++ dataset to the human radiance fields learned on the DeepFashion dataset in \figref{fig:novel_pose} and compare with ENARF. As ENARF uses a learned blend weight field, it produces artifacts at the joints given such out-of-distribution poses. We further show pose control results of our model trained on Surreal and AIST++ in \figref{fig:surreal}a and \figref{fig:aist}a, respectively. The results suggest that our method achieves convincing pose control results on various datasets.

\boldparagraph{Shape Control} 
\figref{fig:Shape_Controllability} demonstrate our shape controllability, achieved by manipulating the SMPL shape parameters $\btheta$. Note that our method generalizes well to the unseen type of shapes. Similarly, we show shape control results of our models trained on Surreal and AIST++ in \figref{fig:surreal}b and \figref{fig:aist}b.

\boldparagraph{Part Control} 
We demonstrate our part editing results in \figref{fig:Part_Controllability}. 
As can be seen, we can edit the head, the upper body, and the lower body independently, with clothes color, clothes style, hairstyle, and facial changes.
Please refer to our supplementary for more results.


\section{Reconstruction} \label{sec:experiment}

\boldparagraph{Datasets} 
We conduct experiments on ZJU-MoCap dataset~\cite{peng2021neural} to evaluate the effectiveness of our vertex-based representation when supervised by a reconstruction loss. ZJU-MoCap records human motion using 21 synchronized cameras and provides accurate human poses. We use one camera for training and the remaining ones for testing novel view synthesis. We divide the motion sequence frames into ``training pose'' and ``novel pose'', and use the former to train the model and the latter to test the performance under the novel pose.

\boldparagraph{Evaluation Metrics} 
We evaluate PSNR, SSIM~\cite{SSIM}, and LPIPS~\cite{lpips} to compare the quality of synthesized images under novel view and novel pose settings.

\boldparagraph{Baselines} 
We compare our method with a state-of-the-art reconstruction method, HumanNeRF~\cite{Weng2022CVPR}, using learned weight field, to demonstrate our method's representation capability and generalization ability. We use their released model to test for novel poses. We also compare with the ``Tri. + Surf.'' baseline in this reconstruction experiment. 



\begin{figure}[t]
     \setlength{\tabcolsep}{0pt}
     \def\mywidth{.33}
     \begin{tabular}{ccc}
      \includegraphics[width=\mywidth\linewidth]{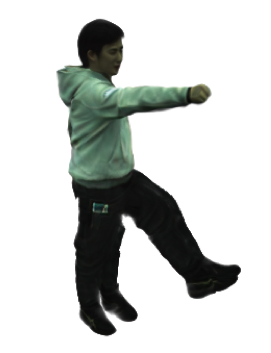} &
      \includegraphics[width=\mywidth\linewidth]{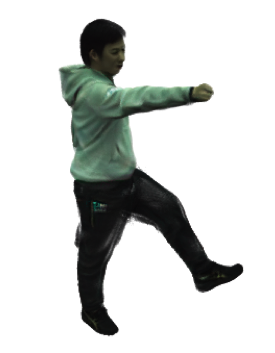} &
      \includegraphics[width=\mywidth\linewidth]{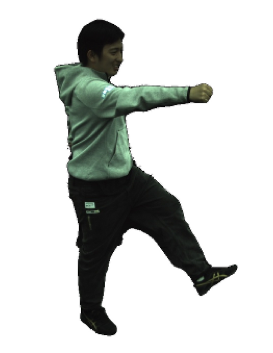} \\
      \includegraphics[width=\mywidth\linewidth]{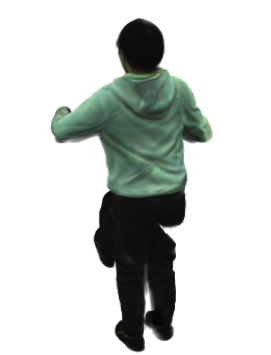} &
      \includegraphics[width=\mywidth\linewidth]{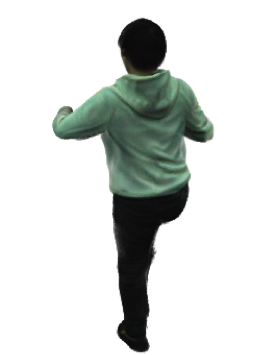} &
      \includegraphics[width=\mywidth\linewidth]{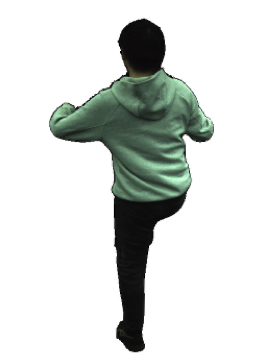} \\
      \begin{small}Tri. + Surf.\end{small} &
      \begin{small}Ours\end{small} &
      \begin{small}GT\end{small} 
     \end{tabular}\vspace{-0.1cm}
     \caption{\textbf{Qualitative Comparison} of the reconstruction experiment. Surface-based mapping suffers from ghosting artifacts, i.e., an extra leg appears in the empty space.}
   \label{fig:ghost}
    \end{figure}


\subsection{Reconstruction Results}
\tabref{tab:ZJU} shows our quantitative comparison of the reconstruction experiments. We compute averaged results over six subjects following the evaluation setting of HumanNeRF. Note that our method achieves competitive performance to the state-of-the-art human reconstruction method HumanNeRF. The tri-plane-based baseline achieves similar results on PSNR and SSIM, but degenerates significantly on LPIPS. This may be caused by ghosting artifacts as shown in \figref{fig:ghost}, \ie, an extra leg appears in the empty space. In contrast, by transforming each 3D point to the local coordinate systems of its nearest neighbors, our method does not suffer from such a problem.

\subsection{Ablation Study}

\boldparagraph{Number of Neighboring Vertices} 
\tabref{tab:ablation_study} shows our quantitative comparison of the reconstruction experiments.
We compare different numbers of neighboring vertices. The performance improves as the number of reference neighbors increases to K=3 but then saturates with K=5. Therefore, we adopt K=3 as the default setting in our main experiments.

\begin{table}[h]
  \centering
  \footnotesize
  \setlength{\tabcolsep}{3pt}
  \begin{tabular}{c|ccc|ccc}
    \toprule
     & \multicolumn{3}{c|}{Novel View} & \multicolumn{3}{c}{Novel Pose}\\
    Method & PSNR~$\uparrow$ & SSIM~$\uparrow$ & LPIPS~$\downarrow$ & PSNR~$\uparrow$ & SSIM~$\uparrow$ & LPIPS~$\downarrow$\\
    \midrule
    HumanNeRF& \textbf{30.24} & 0.9680 & 31.73 & 29.68 & 0.9664 &32.87 \\
    Tri. + Surf. & 29.21 & 0.9631 &36.50 & 29.96 & 0.9678& 31.44 \\
    Ours & 30.11 & \textbf{0.9684} & \textbf{30.24} & \textbf{30.25} & \textbf{0.9698} & \textbf{28.53} \\
    \bottomrule
  \end{tabular}
  \vspace{-0.2cm}
  \caption{\textbf{Reconstruction Results} on ZJU-MoCap.}
  \label{tab:ZJU}
\end{table}

% ...
\vspace{0.35cm}
\begin{table}[h]
 \footnotesize
  \centering
  \setlength{\tabcolsep}{3pt}
  \begin{tabular}{c|ccc|ccc}
    \toprule
     & \multicolumn{3}{c}{Novel View} & \multicolumn{3}{|c}{Novel Pose}\\
    Method & PSNR~$\uparrow$ & SSIM~$\uparrow$ & LPIPS~$\downarrow$ & PSNR~$\uparrow$ & SSIM~$\uparrow$ & LPIPS~$\downarrow$\\
    \midrule
    K=1 & 30.03 & 0.9683 & 30.59 & 30.13 & 0.9694 & 28.84\\
    K=3 & \textbf{30.11} & \textbf{0.9684} & 30.24 & \textbf{30.25} & \textbf{0.9698} & 28.53 \\
    K=5 & 30.05 & 0.9684 & \textbf{30.09} & 30.19 & 0.9697 & \textbf{28.41}\\
    w/o direction & 29.72 & 0.9652 & 33.43 & 29.97 & 0.9673& 30.92  \\
    \bottomrule
  \end{tabular}
  \caption{\textbf{Ablation Study} on ZJU-MoCap.}
  \label{tab:ablation_study}
  \vspace{-0.2cm}
\end{table}

\boldparagraph{Direction in Local Coordinate System} 
In the last row of \tabref{tab:ablation_study},
we remove the direction of sampled points in the local coordinate system. It means the sampled points with the same distances to the neighboring vertex have the same representation, thus leading to worse performance. This experimental result verifies our design choice.
\section{Conclusions}We present VeRi3D, a generative vertex-based human radiance field that offers versatile controllability over camera pose, human pose, shape, and part-level editing. We demonstrate that our vertex-based human representation achieves superior performance on unconditional human image synthesis and is also competitive in the reconstruction setting. 

\noindent{\bf Limitations:} Our method relies on accurate pose distribution to be able to respect the input pose. With mismatched pose distribution, \eg, bent legs estimated as straight ones, our method needs to generate bent legs given straight legs as input, thus not respecting the input pose. This could be addressed by jointly refining the pose distribution in future work. Further, our method may have artifacts in less observed regions, \eg, there are sometimes artifacts in the back regions for forward-facing datasets.  This may be resolved by adding regularizations to unobservable regions. 

\subsection*{Acknowledgements}
This work was supported by NSFC under grant U21B2004, 62202418, the Fundamental Research Funds for the Central Universities under Grant No. 226-2022-00145, and the Zhejiang University Education Foundation Qizhen Scholar Foundation. We thank Sheng Miao for proofreading the manuscript and Yue Wang for helpful discussions.

{\small
\bibliographystyle{ieee_fullname}
\bibliography{bibliography_long, bibliography, bibliography_custom}
}

\newpage
\onecolumn

\begin{center}
\Large{\textbf{Supplementary Material for VeRi3D: Generative Vertex-based Radiance Fields for 3D Controllable Human Image Synthesis}} 
\end{center}
\appendix

\section{Implementation Details}

\subsection{VeRi3D}

\boldparagraph{Network Architecture} 
%
The generator composes of three parts: a mapping network, a convolutional backbone, and an MLP decoder.
The mapping network and the convolutional backbone follow the official implementation\footnote{\url{https://github.com/NVlabs/stylegan3}} of StyleGAN2~\cite{Karras2019stylegan2}. The convolutional backbone consists of 7 style blocks, conditioning on a 256-dimensional Gaussian noise input and producing a 64-channel 256 ${\times}$ 256 feature image. Each block consists of modulation, convolution, and normalization. 
The MLP decoder consists of 2 hidden layers of 256 units. The input to the MLP includes the 64-channel averaged feature vector $\bar{\bff}$ and the positional encoded local coordinate information $\gamma({\bx^l})$ with $L=10$.

\subsection{Datasets} 
\boldparagraph{Surreal} 
Following ENARF~\cite{Noguchi2022ECCV}, we crop the first frame of all videos to 180 ${\times}$ 180 with the center at the pelvis joint and then resize it to 128 ${\times}$ 128. 68033 images are obtained in total. The background mask provided by the dataset is used to replace the background with black color. 

\boldparagraph{AIST++} 
Following ENARF~\cite{Noguchi2022ECCV}, we crop the images to 600 ${\times}$ 600 with the center at the pelvis joint, and then resize it to 256 ${\times}$ 256. 3000 frames are sampled for each subject, resulting in 90K frames in total. We use an off-the-shelf segmentation model~\cite{paddleseg2019} to paint the background black. 

\boldparagraph{DeepFashion} 
We use the data provided by~\cite{hong2023evad}. The dataset filters out images with partial observations and inaccurate SMPL estimations, resulting in 8K images for training. We render the images at the resolution of 512 ${\times}$ 256 pixels.

\section{Additional Experimental Results}
\label{sec:additional_results}

\subsection{Additional Qualitative Results}
\begin{figure*}[h]
    \centering
     \setlength{\tabcolsep}{0pt}
     \def\mywidth{.48}
     \begin{tabular}{cc}
      \includegraphics[width=\mywidth\linewidth]{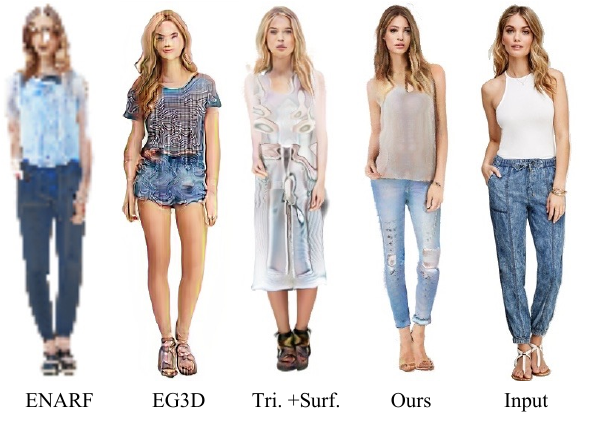} &
      \includegraphics[width=\mywidth\linewidth]{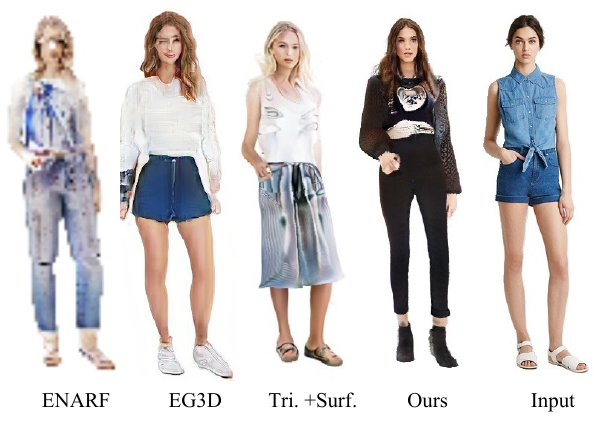}\\
     \end{tabular}\vspace{-0.4cm}
     \caption{\textbf{Qualitative Comparison on DeepFashion.} We provide the corresponding real image as a reference for each input pose.}
   \label{fig:pose condition}
    \end{figure*}

\boldparagraph{DeepFashion}
 \figref{fig:pose condition} shows a qualitative comparison on the DeepFashion dataset. We provide the corresponding real image as a reference for each input pose. Our method has more fidelity and higher pose accuracy.

\begin{figure*}[h]
  \centering
      \includegraphics[width=\linewidth]{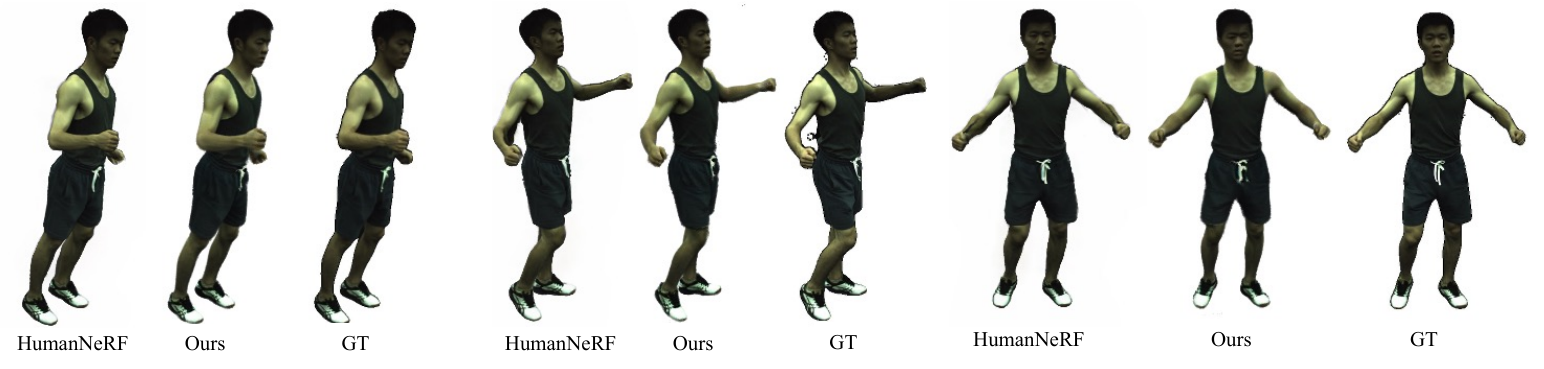}
     \caption{\textbf{Qualitative Comparison with HumanNerf on ZJU-MoCap.}}
   \label{fig:compare_with_humannerf}
    \end{figure*}

\boldparagraph{ZJU-MoCap}
 \figref{fig:compare_with_humannerf} shows a qualitative comparison with HumanNerf on ZJU-MoCap. Our method achieves competitive performance to the state-of-the-art human reconstruction method HumanNeRF.

\begin{figure*}[h]
    \centering
    \vspace{-0.2cm}
     \setlength{\tabcolsep}{0pt}
     \def\mywidth{5cm}
     \begin{tabular}{P{0.5cm}P{5.5cm}P{5.5cm}P{5.5cm}}
      \rotatebox{90}{Head Control} & \includegraphics[width=\mywidth]{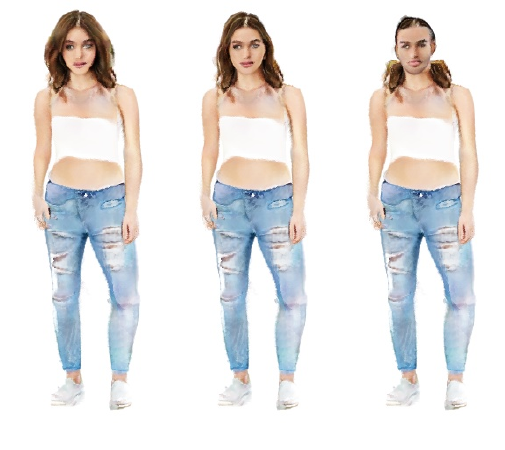} &
      \includegraphics[width=\mywidth]{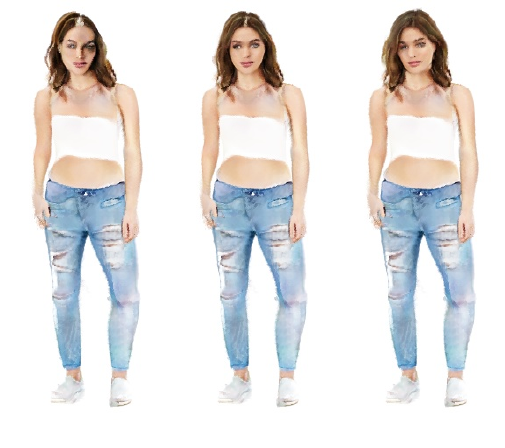} &
      \includegraphics[width=\mywidth]{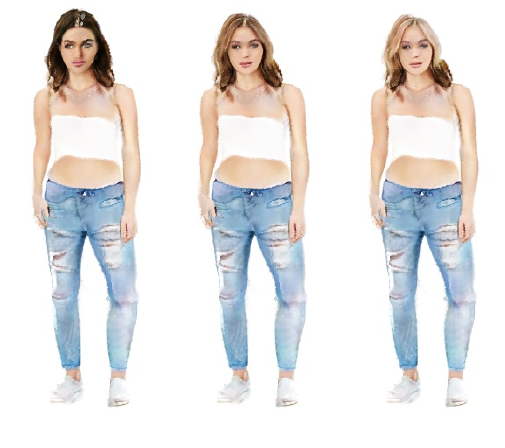} \\
      \vspace{-0.1cm}
      \rotatebox{90}{Upper Body Control} &\includegraphics[width=\mywidth]{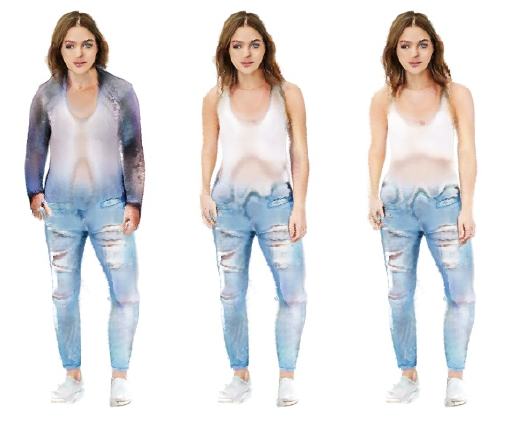} &
      \includegraphics[width=\mywidth]{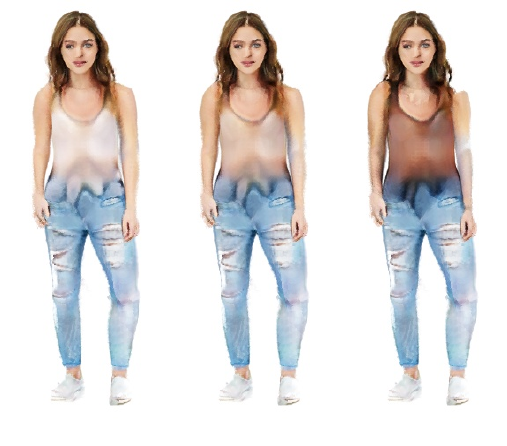} &
      \includegraphics[width=\mywidth]{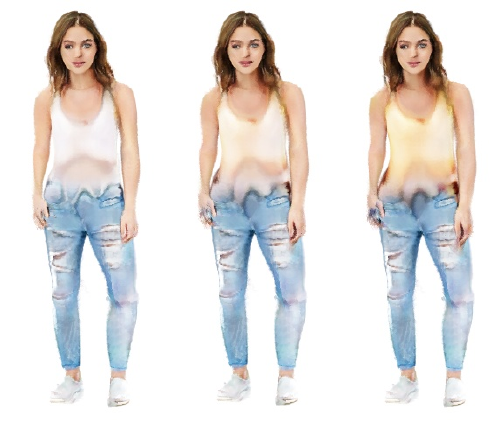} \\
      \vspace{-0.1cm}
      \rotatebox{90}{Lower Body Control} &\includegraphics[width=\mywidth]{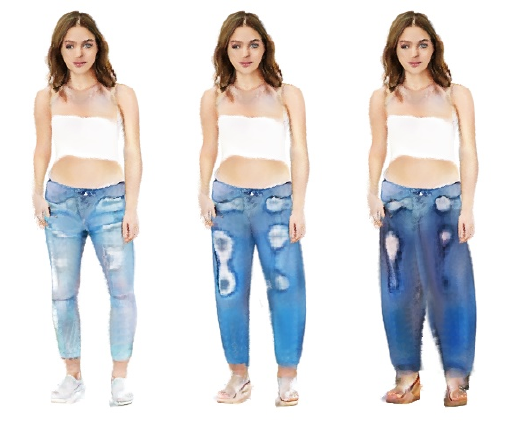} &
      \includegraphics[width=\mywidth]{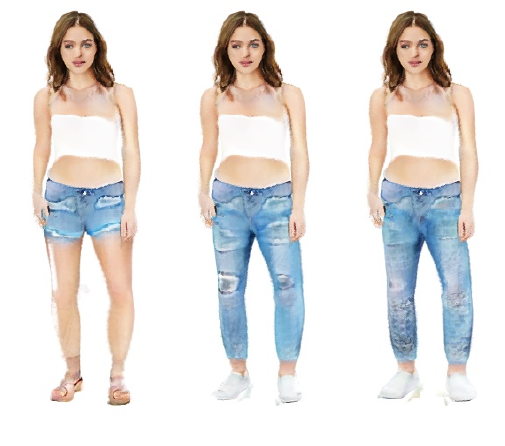} &
      \includegraphics[width=\mywidth]{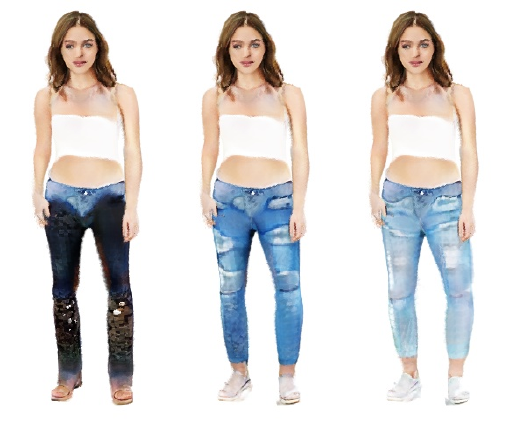} \\
     \end{tabular}\vspace{-0.1cm}
     \caption{\textbf{Additional Qualitative Results} for different PCA Components. We show three components for each part and three samples for each component.}
   \label{fig:part_control}
    \end{figure*}

\subsection{Part Control}
 \boldparagraph{Part Control using Different PCA Components}
We further show the results of different PCA components in \figref{fig:part_control}. Here, we change one component coefficient for each part respectively and keep the remaining component coefficients fixed. We find that different components capture different semantic meanings. For example, the components of the head capture hair length and color, whereas the components of the upper/lower body control cloth length, tightness, and color. 
Interestingly, we observe that one component of the head controls its orientation. We note that this is due to the distribution mismatch between the GT poses and the noisy SMPL poses, where the generator needs to model the orientation to compensate for the inaccurate pose distribution.

\section{Limitations}
Our method may be negatively affected by inaccurate data. In the DeepFashion dataset, the distribution of the estimated SMPL poses mismatches the image poses. Many bent legs are estimated as straight ones. To match the image pose distribution during training, the model needs to generate the bent leg on top of the straight SMPL leg as a sort of ``clothes''. Therefore changing the appearance may alter the pose. This could be addressed by jointly refining the pose distribution in future work. In AIST++ dataset, the background is not removed accurately, especially in the hand regions. Therefore our model may also generate stuff around the hand to match the image distribution.

\end{document}


\title{Supplementary Material for \\ VeRi3D: Generative Vertex-based Radiance Fields for \\ 3D Controllable Human Image Synthesis}

\author{
Xinya Chen$^{1}$,
~Jiaxin Huang$^{1}$,
~Yanrui Bin$^{2}$,
~Lu Yu$^{1}$,
~Yiyi Liao$^{1}$\thanks{Corresponding author.} 
\vspace{0.8em}\\
$^{1}$Zhejiang University~$^{2}$Huazhong University of Science and Technology\\
{\tt\small \{hust.xinyachen,~jaceyh919,~binyanrui\}@gmail.com, \{yul,~yiyi.liao\}@zju.edu.cn}
}
\maketitle

\begin{abstract}
    In this \textbf{supplementary document}, we first clarify implementation details in Section 1.
    Next, we provide more results for part control and additional experimental results in Section 2.
    Finally, we discuss our limitations in Section 3.
\end{abstract}

\section{Implementation Details}
In this section, we describe the implementation details of our proposed VeRi3D. We also provide more details of the datasets.

\subsection{VeRi3D}

\boldparagraph{Network Architecture} 
%
The generator composes of three parts: a mapping network, a convolutional backbone, and an MLP decoder.
The mapping network and the convolutional backbone follow the official implementation\footnote{\url{https://github.com/NVlabs/stylegan3}} of StyleGAN2~\cite{Karras2019stylegan2}. The convolutional backbone consists of 7 style blocks, conditioning on a 256-dimensional Gaussian noise input and producing a 64-channel 256 ${\times}$ 256 feature image. Each block consists of modulation, convolution, and normalization. 
The MLP decoder consists of 2 hidden layers of 256 units. The input to the MLP includes the 64-channel averaged feature vector $\bar{\bff}$ and the positional encoded local coordinate information $\gamma({\bx^l})$ with $L=10$.

\subsection{Datasets} 
\boldparagraph{Surreal} 
Following ENARF~\cite{Noguchi2022ECCV}, we crop the first frame of all videos to 180 ${\times}$ 180 with the center at the pelvis joint and then resize it to 128 ${\times}$ 128. 68033 images are obtained in total. The background mask provided by the dataset is used to replace the background with black color. 

\boldparagraph{AIST++} 
Following ENARF~\cite{Noguchi2022ECCV}, we crop the images to 600 ${\times}$ 600 with the center at the pelvis joint, and then resize it to 256 ${\times}$ 256. 3000 frames are sampled for each subject, resulting in 90K frames in total. We use an off-the-shelf segmentation model~\cite{paddleseg2019} to paint the background black. 

\boldparagraph{DeepFashion} 
We use the data provided by~\cite{hong2023evad}. The dataset filters out images with partial observations and inaccurate SMPL estimations, resulting in 8K images for training. We render the images at the resolution of 512 ${\times}$ 256 pixels.

\section{Additional Experimental Results}
\label{sec:additional_results}

\subsection{Part Control}

\begin{figure*}[h]
    \centering
    \vspace{-0.2cm}
     \setlength{\tabcolsep}{0pt}
     \def\mywidth{5cm}
     \begin{tabular}{P{0.5cm}P{5.5cm}P{5.5cm}P{5.5cm}}
      \rotatebox{90}{Head Control} & \includegraphics[width=\mywidth]{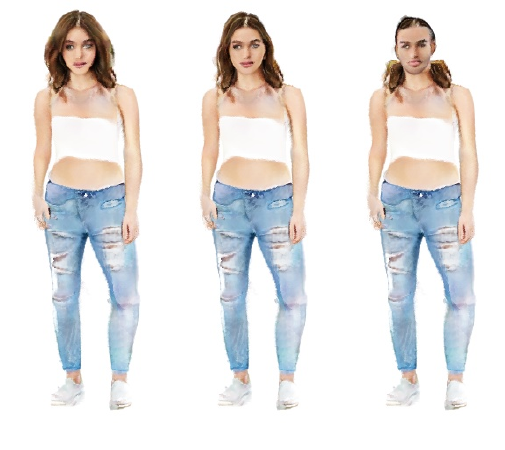} &
      \includegraphics[width=\mywidth]{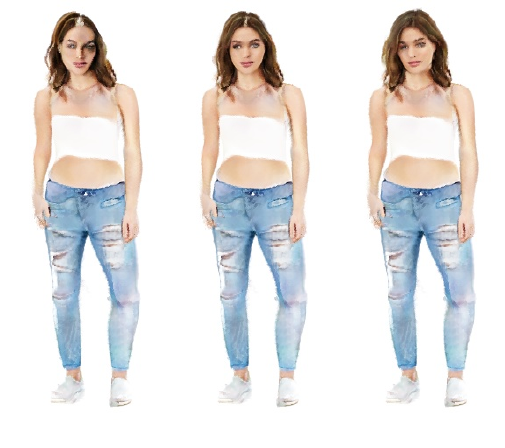} &
      \includegraphics[width=\mywidth]{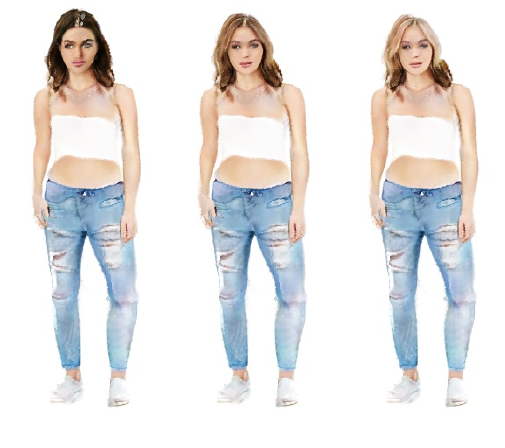} \\
      \vspace{-0.1cm}
      \rotatebox{90}{Upper Body Control} &\includegraphics[width=\mywidth]{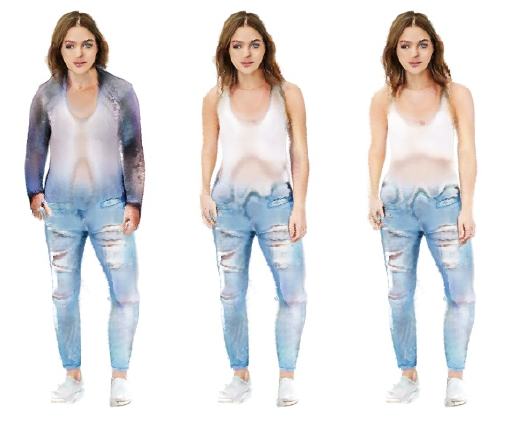} &
      \includegraphics[width=\mywidth]{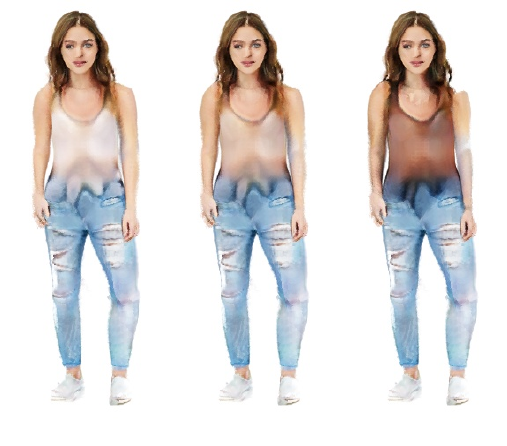} &
      \includegraphics[width=\mywidth]{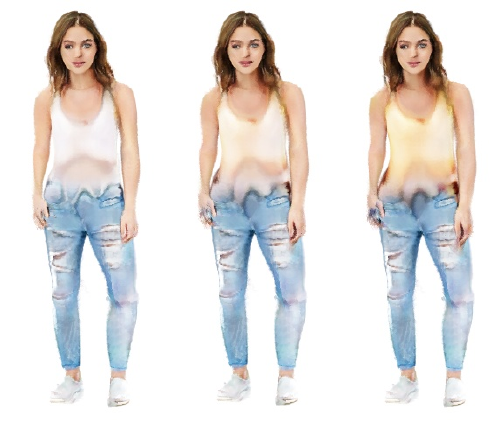} \\
      \vspace{-0.1cm}
      \rotatebox{90}{Lower Body Control} &\includegraphics[width=\mywidth]{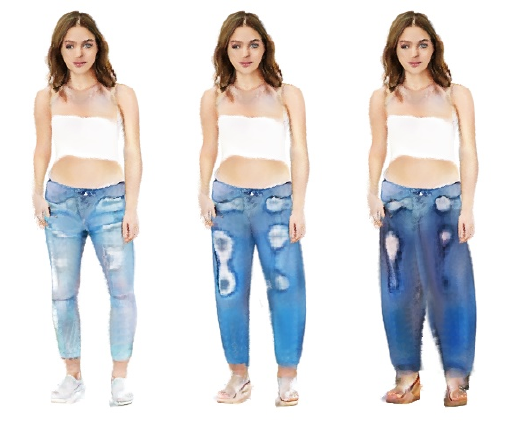} &
      \includegraphics[width=\mywidth]{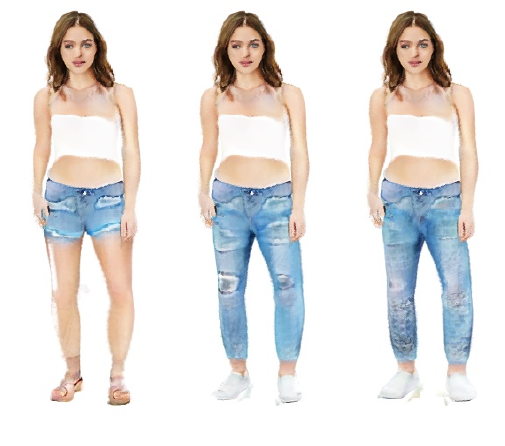} &
      \includegraphics[width=\mywidth]{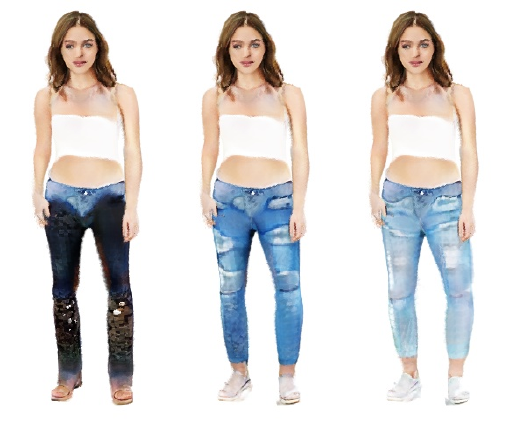} \\
     \end{tabular}\vspace{-0.1cm}
     \caption{\textbf{Additional Qualitative Results} for different PCA Components. We show three components for each part and three samples for each component.}
   \label{fig:part_control}
    \end{figure*}

\boldparagraph{Part Control using Different PCA Components}
We further show the results of different PCA components in \figref{fig:part_control}. Here, we change one component coefficient for each part respectively and keep the remaining component coefficients fixed. We find that different components capture different semantic meanings. For example, the components of the head capture hair length and color, whereas the components of the upper/lower body control cloth length, tightness, and color. 
Interestingly, we observe that one component of the head controls its orientation. We note that this is due to the distribution mismatch between the GT poses and the noisy SMPL poses, where the generator needs to model the orientation to compensate for the inaccurate pose distribution.

\subsection{Additional Qualitative Results}

\begin{figure}[h]
     \setlength{\tabcolsep}{0pt}
     \def\mywidth{.48}
     \begin{tabular}{cc}
      \includegraphics[width=\mywidth\linewidth]{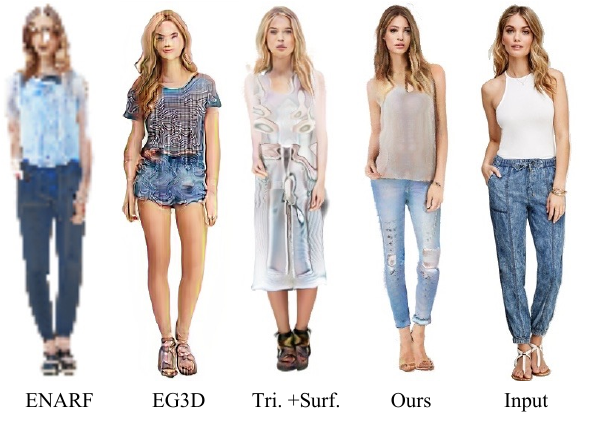} &
      \includegraphics[width=\mywidth\linewidth]{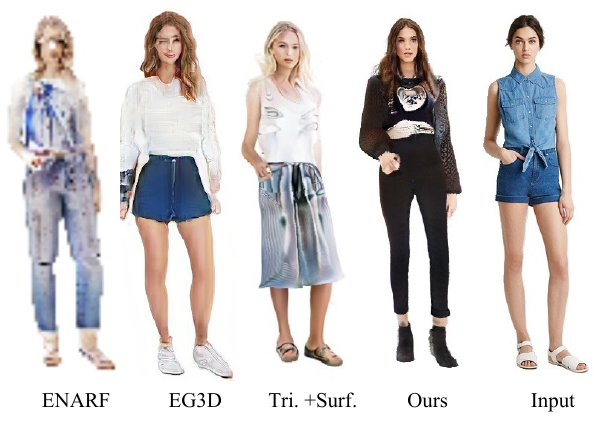}\\
     \end{tabular}\vspace{-0.4cm}
     \caption{\textbf{Qualitative Comparison on DeepFashion.} We provide the corresponding real image as a reference for each input pose.}
   \label{fig:pose condition}
    \end{figure}

\begin{figure}[h]
     \setlength{\tabcolsep}{0pt}
     \def\mywidth{1}
     \begin{tabular}{c}
      \includegraphics[width=\mywidth\linewidth]{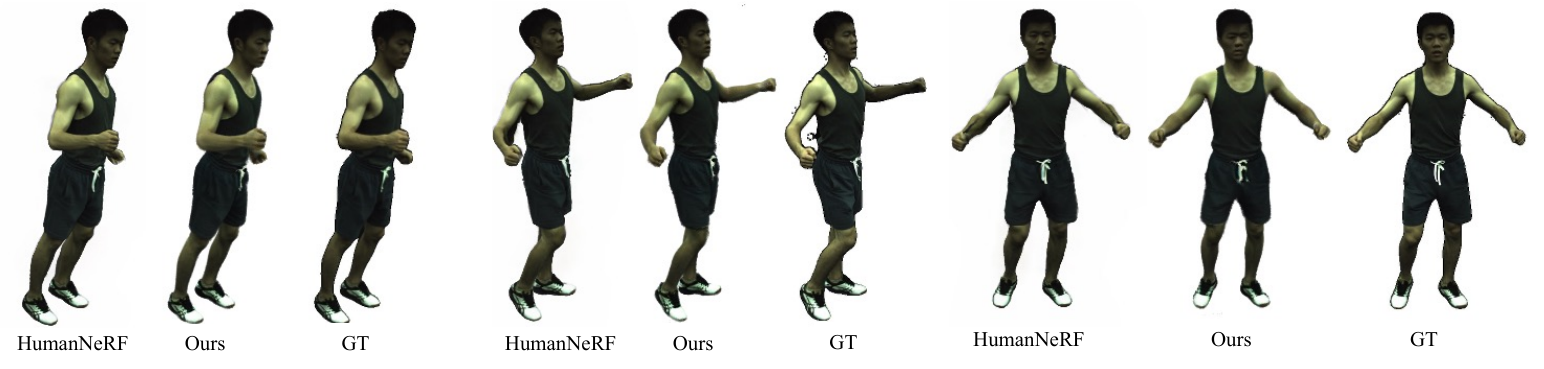}\\
     \end{tabular}\vspace{-0.4cm}
     \caption{\textbf{Qualitative Comparison with HumanNerf on ZJU-MoCap.}}
   \label{fig:compare_with_humannerf}
    \end{figure}
    
    \vspace{-0.4cm}

\boldparagraph{DeepFashion}
 \figref{fig:pose condition} shows a qualitative comparison on the DeepFashion dataset. We provide the corresponding real image as a reference for each input pose. Our method has more fidelity and higher pose accuracy.

 \boldparagraph{ZJU-MoCap}
 \figref{fig:compare_with_humannerf} shows a qualitative comparison with HumanNerf on ZJU-MoCap. Our method achieves competitive performance to the state-of-the-art human reconstruction method HumanNeRF.

\section{Limitations}
Our method may be negatively affected by inaccurate data. In the DeepFashion dataset, the distribution of the estimated SMPL poses mismatches the image poses. Many bent legs are estimated as straight ones. To match the image pose distribution during training, the model needs to generate the bent leg on top of the straight SMPL leg as a sort of ``clothes''. Therefore changing the appearance may alter the pose. This could be addressed by jointly refining the pose distribution in future work. In AIST++ dataset, the background is not removed accurately, especially in the hand regions. Therefore our model may also generate stuff around the hand to match the image distribution.




{\small
\bibliographystyle{ieee_fullname}
\bibliography{bibliography_long, bibliography, bibliography_custom}
}